\documentclass[lettersize,journal]{IEEEtran}
\usepackage{amsmath,amsfonts}
\usepackage{algorithmic}
\usepackage{algorithm}
\usepackage{array}
\usepackage[caption=false,font=normalsize,labelfont=sf,textfont=sf]{subfig}
\usepackage{textcomp}
\usepackage{stfloats}
\usepackage{url}
\usepackage{verbatim}
\usepackage{graphicx}
\usepackage{cite}
\usepackage{color}
\begin{document}

\title{A Hybrid End-to-End Spatio-Temporal Attention Neural Network with Graph-Smooth Signals for EEG Emotion Recognition}

\author{Shadi Sartipi,~\IEEEmembership{Student Member,~IEEE,} Mastaneh Torkamani-Azar,~\IEEEmembership{Member,~IEEE,} and Mujdat Cetin,~\IEEEmembership{Fellow,~IEEE}
% \IEEEmembership{Staff,~IEEE,}
        % <-this % stops a space
\thanks{This work has been partially supported by the National Science Foundation (NSF) under grants CCF-1934962 and DGE-1922591. ($corresponding~author:~Shadi~Sartipi$)}% <-this % stops a space
\thanks{Shadi Sartipi and Mujdat Cetin are with the Department of Electrical and Computer Engineering, University of Rochester, Rochester, NY 14627, USA ({ssartipi@ur.rochester.edu, mujdat.cetin@rochester.edu}).}
\thanks{Mastaneh Torkamani-Azar is with the A. I. Virtanen Institute for Molecular Sciences, University of Eastern Finland, 70211 Kuopio, Finland ({mastaneh.torkamani@uef.fi}).}%
\thanks{Mujdat Cetin is with the Goergen Institute for Data Science, University of Rochester, Rochester, NY 14627, USA.}}

% The paper headers
\markboth{IEEE Transactions on Cognitive and Developmental Systems}%
{Sartipi \MakeLowercase{\textit{et al.}}: A Hybrid Neural Network for EEG Emotion Recognition}

% \IEEEpubid{0000--0000/00\$00.00~\copyright~2021 IEEE}
% Remember, if you use this you must call \IEEEpubidadjcol in the second
% column for its text to clear the IEEEpubid mark.

\maketitle

\begin{abstract}
Recently, physiological data such as electroencephalography (EEG) signals have attracted significant attention in affective computing. In this context, the main goal is to design an automated model that can assess emotional states. Lately, deep neural networks have shown promising performance in emotion recognition tasks. However, designing a deep architecture that can extract practical information from raw data is still a challenge. Here, we introduce a deep neural network that acquires interpretable physiological representations by a hybrid structure of spatio-temporal encoding and recurrent attention network blocks. Furthermore, a preprocessing step is applied to the raw data using graph signal processing tools to perform graph smoothing in the spatial domain. We demonstrate that our proposed architecture exceeds state-of-the-art results for emotion classification on the publicly available DEAP dataset. To explore the generality of the learned model, we also evaluate the performance of our architecture towards transfer learning (TL) by transferring the model parameters from a specific source to other target domains. Using DEAP as the source dataset, we demonstrate the effectiveness of our model in performing cross-modality TL and improving emotion classification accuracy on DREAMER and the Emotional English Word (EEWD) datasets, which involve EEG-based emotion classification tasks with different stimuli. 
\end{abstract}


\section{Introduction}
\IEEEPARstart{A}{ffective} computing is a popular field of study wherein researchers try to develop automatic recognition systems or devices that can interpret or respond to human emotional states. Brain-Computer Interfaces (BCI) link brain activity with external devices \cite{wolpaw2002brain}. Recently, emotion recognition using physiological signals attracted a notable amount of attention \cite{gannouni2021emotion}. Physiological signals acquired with wearable devices include electroencephalogram (EEG), electrocardiogram (ECG), electromyogram (EMG), blood pressure, galvanic skin response (GSR), eye-tracking metrics such as pupil dilation and gaze entropy, body temperature, and movement kinematics, to name a few. The neural activity of cortical regions can be recorded by multichannel EEG in a way that can preserve the spectral and rhythmic characteristics of brain signals \cite{alarcao2017emotions}. Comparing EEG with other non-invasive recording methods shows that EEG has better temporal resolution and can acquire brain signals per millisecond. High temporal resolution and ease of use made EEG one of the most practical ways of handling tasks related to cognitive and affective reactions \cite{alarcao2017emotions}. However, EEG suffers from low signal-to-noise ratio (SNR) and poor spatial resolution. Accordingly, it is challenging to use EEG signals for downstream tasks when compared to MRI and fNIRS \cite{nguyen2017utilization, al2021bimodal}.

EEG signals are primarily analyzed in particular frequency bands, including theta ($\theta: 4$-$8$ Hz), alpha ($\alpha: 8$-$12$ Hz), beta ($\beta: 12$-$29$ Hz), and gamma ($\gamma:~>30$ Hz). Most of the early pieces of work on EEG-based emotion classification have generally relied on two main steps: extracting the informative features and defining a supervised machine learning approach. Wang \emph{et~al.} evaluated the performance of three different features, namely power spectral density (PSD), wavelet entropy, and nonlinear dynamical features with kernel support vector machine (SVM) \cite{wang2014emotional}. Zheng \emph{et~al.} \cite{zheng2015investigating} investigated the critical frequency bands and channels with differential entropy (DE), DE asymmetry, and PSD features. They explored the performance of different features by \emph{K}-nearest neighbors (\emph{K}-NN), SVM, and deep belief networks (DBN). In \cite{bhattacharyya2020novel}, the authors offered an approach to calculate spectral and temporal entropies by decomposing EEG data via Fourier-Bessel series expansion-based empirical wavelet transform. Then, \emph{K}-NN and Shannon entropies were computed after multi-scaling operation in the spectral and temporal domains. \cite{li2022emotion} proposed a new rhythm sequencing approach to find the best rhythmic features from the sequence of multi-channel EEG data. Finally, \cite{yang2015takagi} applied transfer learning to address the issue of distribution shift between the training and test data.

Nevertheless, covering all the manually extracted features both in time and frequency domains is complicated. Besides, the susceptibility of EEG signals to artifacts severely degrades the performance of classical machine learning approaches \cite{jafarifarmand2019eeg}. To address this issue, one can exploit end-to-end models that start with raw signals rather than extracted features. End-to-end deep learning (DL) approaches have been indicated to surpass classic approaches in various fields, including speech recognition \cite{deng2013new}, computer vision \cite{jaderberg2015spatial}, and biomedical signal processing \cite{goh2018spatio}. In contrast to shallow classifiers that require feature engineering, deep neural networks automatically extract practical features from the given signals and learn the low- and high-level representations of the input data \cite{lecun2015deep}. Several deep learning architectures and methodologies have been proposed for EEG-based BCIs; see e.g.,~\cite{zhang2020survey,tao2020eeg, torkamani2020prediction,huang2020s}. Schirrmeister \emph{et~al.}\cite{schirrmeister2017deep} designed a deep convolutional neural network (CNN) based architecture with temporal and spatial convolutional (conv) filters followed by conv-pooling blocks to reduce the dimension. Zhao \emph{et~al.} investigated the fusion of three different modalities, namely EEG, raw eye-movement-images (EIG), and eye movement features (EYE) \cite{zhao2021multimodal}. Feeding the fusion of different modalities to a dense co-attention symmetric network resulted in higher classification performance than the single modality. In \cite{tao2020eeg}, authors applied an attention technique to set different weight importance to EEG channels. Then, they extracted spatio-temporal features by applying CNN and a recurrent neural network (RNN). Multi-column convolutional neural network (MCNN) was introduced by authors of \cite{yang2019multi} for emotional states classification. They tested their work in a subject-independent manner by assuming five participants as the test data without performing cross-validation. A novel architecture coined as frame-level distilling neural network (FLDNet), that learns the distilled properties from the correlation of various frames, was introduced in \cite{wang2021fldnet}. They presented a triple-net structure to distill the learned features of each net consecutively. The deep forest was proposed in \cite{cheng2020emotion}, where EEG data were converted to two-dimensional (2D) frame sequences by considering the spatial position of the EEG channels and then fed to the model. The proposed model was insensitive to hyper-parameter settings.
% Although DL models have been successful in EEG analysis, there are few works that explore the learned representation generalization ability on another dataset with a similar task \cite{siddharth2019utilizing}.
EEG-based emotion recognition via DL is still in its early years. Therefore, there is yet to find a better deep structure. Furthermore, although DL models have been successful in EEG analysis, there are few studies that not only explore the performance of the learned representation but also consider their generalization ability on another dataset with a similar task \cite{siddharth2019utilizing}.  

The main goal of this study is to design a new deep architecture to enhance the performance of the current algorithms for EEG-based emotion recognition. It has been established in different areas that CNNs are effective in capturing the spatial representations, while RNNs capture the temporal dependencies well \cite{chen2020emotion}. Recording EEG data from multiple electrodes over the scalp during a period of time forms both spatial and temporal structures. In order to analyze these structured time series successfully, the extracted information from spatial structures and temporal dynamics should be accounted for \cite{shadiembc2021}. We propose the hybrid end-to-end spatio-temporal attention neural network (STANN) with smooth signals over graphs to consider these two aspects of the data within a unified architecture. As the central contribution of this work, STANN consists of two parallel blocks: the spatio-temporal encoding block and the recurrent attention network block. 
% STANN also utilizes the advantage of time scaling as offered by bidirectional attention networks. 
% Concentration of the attention mechanism on specific time scales - by multiplication of hidden state outputs by trainable weights - can be physiologically realized by language-related components of event-related potentials (ERPs) and the time it takes for the brain to perceive and react to emotionally-loaded stimuli. 
Considering the complex structure of brain signals and their time-varying character, we propose the idea of applying the graph Fourier transform (GFT) and low pass graph filtering \cite{saboksayr2021online} in a pre-processing step. GFT is considered as a solution for overcoming the low SNR of EEG signals \cite{higashi2014smoothing}. 
% EEG data related to GFT low frequencies have similar the behavior among neighbor electrodes. 
% This step is called smoothing over the defined graph. 
Accordingly, our proposed smooth signal spatio-temporal attention neural network (SS-STANN) simultaneously learns both spatial information and discriminative time dependencies. To evaluate the performance, we apply the introduced method on the publicly available EEG dataset named DEAP that contains discrete ratings for valence, arousal, and dominance in response to audio-visual stimuli \cite{koelstra2011deap}. In our comprehensive experimental analysis, we demonstrate the superiority of STANN when using either raw EEG signals or smoothed graph signals as an input.

Deep learning models commonly require a large number of parameters to be trained compared to traditional machine learning methods. Thus, DL models need a significant amount of data \cite{zhang2021adaptive}. One of the challenges related to EEG tasks and DL is insufficient labeled data for similar tasks. A number of transfer-based approaches have been proposed to address this issue that leverage a pre-existing large enough dataset known as the source dataset. Yet, there would be inconsistency across target and source domains which necessitates fine-tuning \cite{wronkiewicz2015leveraging} the network for the target data. We demonstrate the applicability of the information learned through our proposed hybrid architecture in similar emotion classification tasks by applying transfer learning and fine-tuning in order to investigate the generality of our proposed model. We consider $5$ layouts for tuning the pre-existing network with confined EEG data. As the target data, we investigate the EEG signals of publicly available DREAMER dataset \cite{katsigiannis2017dreamer} and an Emotional English Word dataset (EEWD) recorded at Sabanci University with different stimuli types that can enlighten the capability of proposed model in cross-modal emotion learning \cite{zhang2021combining}. The major contributions of our work are summarized as follows:
\begin{itemize}
    \item A novel deep architecture that considers spatial and temporal information of time-series data is proposed for EEG emotion classification. The proposed hybrid network encodes the spatio-temporal and attentive temporal information in parallel.
    \item This paper considers the relation among neighbor EEG electrodes to use their spatial-spectral characteristics. Low pass graph filtering is applied to enforce graph smoothness in the spatial domain.
    \item This paper shows the possibility of transferring the learned model parameters for cross-subject and  cross-dataset (similar EEG tasks with different stimuli) scenarios. The results are promising in both scenarios. Improved results indicate the cross-modality capability of the proposed model since the learned representations from EEG signals, elicited in response to video clips, can be used to improve classification accuracy even for datasets of emotional written words.
\end{itemize}

% M: A general comment: the current manuscript and its revision have covered different analysis that will only expand when we want to convince the reviewers. I would like to suggest the following edits: In the intro, use bold sentences to introduce the topic of each paragraph (no need to use section numbers). And here, add section numbers to clearly tell the reader what they will see, or even introduce research questions (see https://ieeexplore.ieee.org/document/9795221). 

A preliminary version of this work was presented in \cite{shadiembc2021}. While \cite{shadiembc2021} contained the initial idea of the STANN framework, this paper extends that preliminary work in several major ways: (1) the approach we present here involves the use of graph signal processing (GSP) tools to perform graph smoothness in the spatial domain, (2) we propose and demonstrate the use of transfer learning within our proposed framework, (3) by visualizing activations of certain layers in our network, we examine how our proposed approach encodes spatial information in the brain, (4) we perform a more extensive comparison of our proposed method to the state-of-the-art, and (5) we extend our experimental analysis to three datasets to demonstrate the effectiveness of the proposed method.

The rest of this paper is structured as follows. Section~\ref{sec:method} introduces the proposed SS-STANN method. Section~\ref{sec:exp} describes the datasets and reports the experimental results. Section~\ref{sec:conclusion} concludes the paper.
\begin{figure*}[!t]
    \centering
    \includegraphics[width=0.9\linewidth]{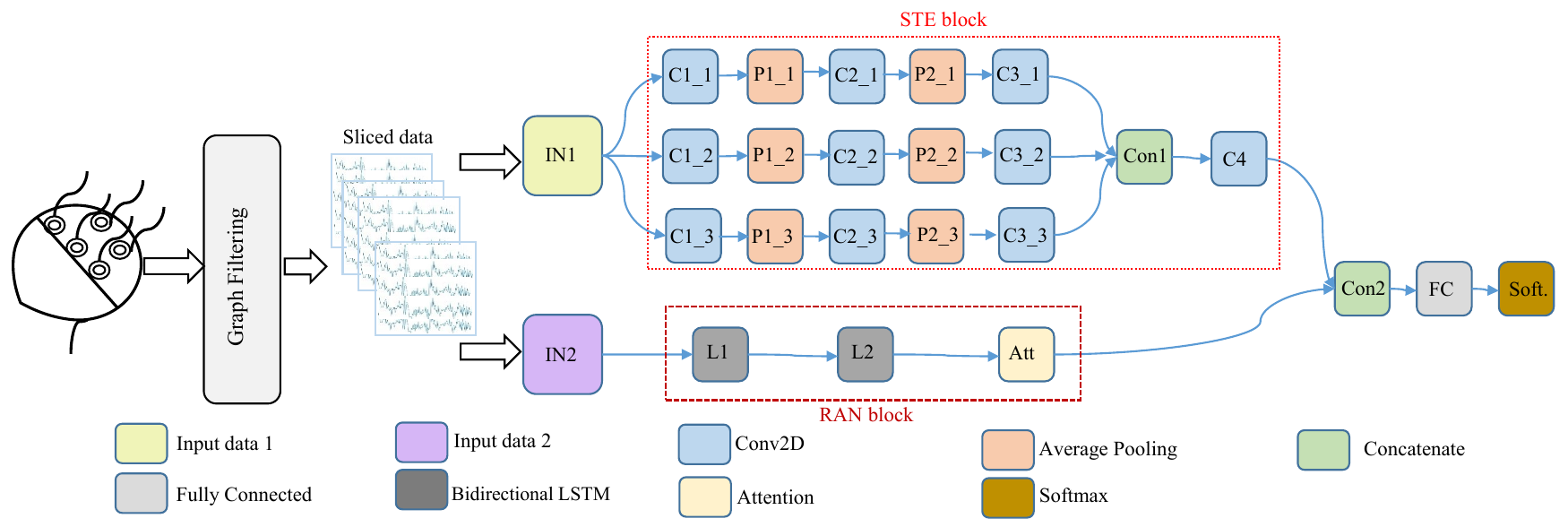}
    \caption{Overview of the proposed SS-STANN architecture. The CNN blocks (blue) contain adopt the ReLU activation function and batch normalization. Pooling blocks (orange) are followed by a dropout layer. IN1 and IN2 correspond to inputs for STE and RAN blocks respectively. (STE: spatio-temporal encoding, RAN: recurrent
attention network).}
    \label{fig1}
\end{figure*}
\section{Method}
\label{sec:method}
\subsection{Overview}
\label{ssec:overview}
Figure \ref{fig1} presents an overview of the proposed pipeline. The EEG data are first graph-smoothed via graph filters, and then sliced by non-overlapping sliding windows to obtain data samples. Next, the data are fed to the proposed deep STANN architecture to learn a discriminative representation for the classification of emotional states. 

\begin{figure}[t]
    \centering
    \begin{minipage}{0.05\textwidth}
\text{\emph{K}$=2$}
\end{minipage}
    \begin{minipage}[c]{.18\textwidth}
    \includegraphics[width=0.8\textwidth]{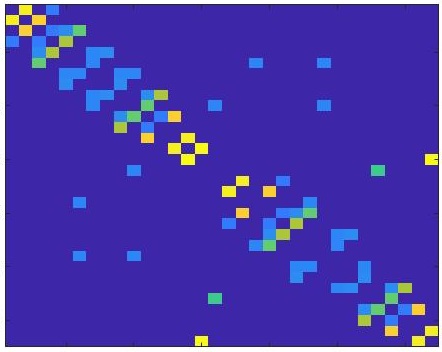}
    \end{minipage}
    \begin{minipage}[c]{.2\textwidth}
    \includegraphics[width=0.8\textwidth]{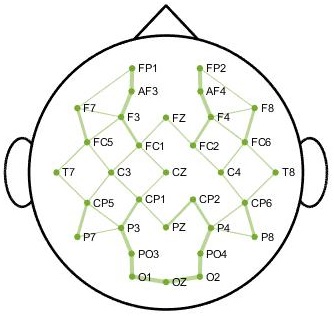}
    \end{minipage}
    \begin{minipage}{0.05\textwidth}
\text{\emph{K}$=4$}
\end{minipage}
    \begin{minipage}[c]{.18\textwidth}
    \includegraphics[width=0.8\textwidth]{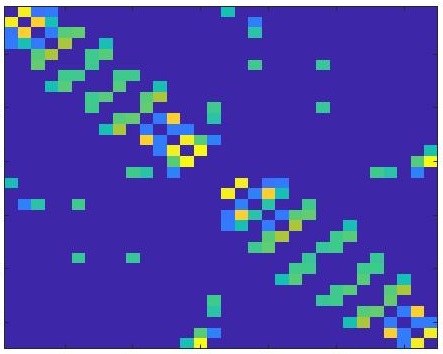}
    \end{minipage}
    \begin{minipage}[c]{.2\textwidth}
    \includegraphics[width=0.8\textwidth]{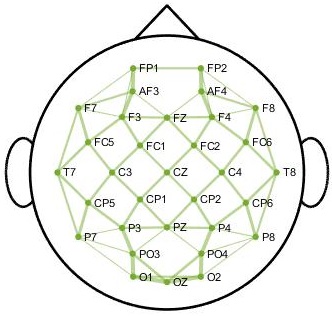}
    \end{minipage}
    
    % \vspace{-0.01in}
    \caption{The adjacency matrix (\textbf{left}) illustration, and its corresponding graph representation (\textbf{right}) from a sample $10$-$20$ electrode placement setup.}
    % \vspace{-0.15in}
    \label{fig2}
\end{figure}
% \begin{figure}[t]
%     \centering
%     \includegraphics[width=0.8\linewidth]{crop_gft_block.pdf}
%     \caption{Graph filtering overview. Adjacency matrix is calculated based on EEG electrodes' positions.}
%     \label{gft}
% \end{figure}
\subsection{Preprocessing using Graph Filtering}
\label{ssec:smooth}

EEG data are recorded during a total time period $T$ from $n$ different electrodes mounted over the scalp, which results in a two-dimensional (2D) signal, $\mathbf{X}\in\mathbb{R}^{n\times{}T}$. Recent works commonly apply hand-crafted features or raw EEG data as the input of deep neural networks. Exploring structural and functional connectivity of the brain \cite{park2013structural} and tracking the relative spatial positions of EEG nodes could be in use of decoding responses elicited from sensory stimuli \cite{li2019eeg}. In order to exploit GSP tools, we need to define an underlying graph. Thus, we calculate the pairwise Euclidean distances of EEG electrodes and build the graph accordingly. In this way, we solely require the Cartesian coordinates of electrodes while the classical common spatial pattern (CSP) filtering depends on individual subjects or tasks. Let us consider an undirected, weighted graph $\mathcal{G}(\mathcal{V},\mathcal{E},\mathbf{A})$, where $\mathcal{V}=\{1,2,...,n\}$ is the set of nodes or EEG channels, $\mathcal{E}\subseteq\mathcal{V}\times\mathcal{V}$ is the set of edges or spatial connections, and $\mathbf{A}\in\mathbb{R}^{n\times{}n}$ is the adjacency matrix. The edge weights $\mathbf{A}_{ij}$ are inversely proportional to the pairwise Euclidean distances between node $i$ and $j$. Subsequently, one can compute the distance $d_{ij}$ as follows:

\begin{equation}
\mathbf{A}_{ij}=d_{ij}^{-1},\;\mathbf{A}_{ii}=0,\;\; \text{for}\; i, j=1,2,...,n.
\end{equation}

For each electrode, \emph{K}-NN is considered to construct the adjacency matrix while keeping it symmetric to represent the brain topology \cite{bullmore2009complex}. To avoid a densely-connected graph, we set \emph{K} to $2$ and $4$. In the literature, a $2$-NN topology was motivated by separating the graph into fronto-temporal and parieto-occipital networks \cite{itani2021graph}. The $4$-NN graph brings engagement with central areas as well. Figure \ref{fig2} shows the final adjacency matrices for these \emph{K} values and their corresponding scalp topologies for $32$ nodes with the $10$-$20$ electrode placement setup.

Furthermore, one can extract informative features using the spectral representation of spatial signals. Using GFT one can analyze the spatial frequency of the signals defined over the graph. The combinatorial graph Laplacian $\mathbf{L}=\mathbf{D}-\mathbf{A}$ is needed for the calculation of GFT in which $$\mathbf{D}_{ii}=\sum_{k}\mathbf{A}_{ik}$$ is a diagonal matrix of nodal degrees \cite{saboksayr2021online}. Given graph signal $\mathbf{X}$ one can compute GFT with respect to $\mathbf{L}$ as follows:
\begin{equation}
\tilde{\mathbf{X}}=\mathbf{V^{T}}\mathbf{X}
\end{equation}
where $\mathbf{V}$ is the orthonormal $n\times{}n$ matrix of eigenvectors of the matrix $\mathbf{L}$.

Since the electrodes installed in adjacent locations detect electrical activities of common sources \cite{higashi2014smoothing}, we apply smoothing via lowpass graph filtering with respect to the defined graph. This will ensure the similarity of the behavior among neighbor electrodes. Low frequencies in the graph correspond to small eigenvalues. Considering $\mathbf{\tilde{h}}=[\tilde{h}_{1}, \tilde{h}_{2}, ..., \tilde{h}_{n}]$ the ideal lowpass filter with bandwidth $w\in\{1,2,...,n\}$ where $\tilde{h}_{i} = 0$ if $i > w$, the GFT coefficients corresponding to the low frequencies with respect to $\mathcal{G}$ are given by:
\begin{equation}
\tilde{\mathbf{X}}_{low}=\mathbf{\tilde{h}}\tilde{\mathbf{X}}
\end{equation}
where $\tilde{h}_i$ is equal to one for $w=[1,\frac{n}{2}]$ and zero otherwise. While we choose the filter bandwidth as $\frac{n}{2}$, users can choose a different number depending on the level of smoothing they want to apply. A smaller bandwidth would lead to more smoothing.

Next, the inverse GFT (iGFT), $\mathbf{X}_{smooth}=\mathbf{V}\mathbf{\tilde{X}}_{low}$, is applied to obtain the smoothed data in the spatial domain \cite{saboksayr2021online}. 
% Figure \ref{gft} presents the steps related to graph filtering process.
\begin{figure}[t]
    \centering
    \includegraphics[width=0.35\textwidth]{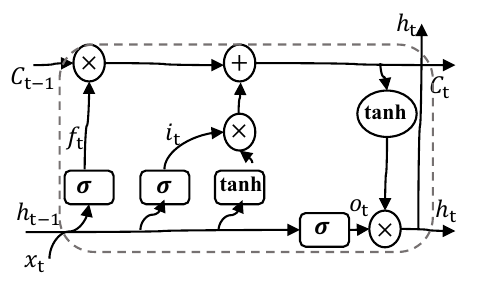}
    \caption{LSTM cell block.}
    \vspace{-0.1in}
    \label{lstmcell}
\end{figure}

\subsection{The Proposed Hybrid SS-STANN Network}
\label{ssec:network}
After the preprocessing and smoothing steps, a sliding window with length $k$ is applied to obtain the EEG slices $i$, i.e., $\mathbf{Z}_i\in\mathbb{R}^{n\times{}k}$. Based on the overview presented in Figure~\ref{fig1}, the proposed network is organized into two main blocks: the spatio-temporal encoding (STE) and the recurrent attention network (RAN) blocks.
% STE block includes MCNN and RAN exploiting the attentive temporal dependencies across different time steps via BiLSTM and attention mechanism
% \subsubsection{Spatio-temporal Encoding}
% \label{sssec:mcnn}

The STE is designed to extract the spatio-temporal information from temporal slices, $\mathbf{Z}_i$. This block consists multiple columns of $2$D CNNs. The STE follows the structure proposed in \cite{ciregan2012, zhang2016single}. This architecture contains several independently acting columns that function as a deep network. All columns receive the same input data and their weights are initialized randomly. The output of this model is equal to the average of all the columns' outputs. In this work, each column contains a series of $2$D CNN, batch normalization, and average-pooling followed by dropout layers with different kernel sizes and a number of filters. Presuming STE has M columns, the feature map of the $m\_th$ column would be $\mathbf{f}^m$. $M$ feature maps,$\{\mathbf{f}^m\}_{m=1}^{m=M}$, are merged to get the final feature map $\mathbf{f}$. The final feature map passes through the $1\times{}1$ conv layer. This structure enables using different kernel sizes per column to detect the informative features at very different temporal and spatial scales across nearby EEG channels. Details of the implementation will be described in Section~\ref{sec:exp}.

% \subsubsection{Recurrent Attention Netweork}
% \label{ssec: lstm}
The RAN block is composed of two bidirectional LSTM layers and an attention mechanism. LSTM networks are recurrent neural networks that capture the dependencies within time steps from sequential data. In LSTMs, the temporal dynamic behavior is captured by feedback connections \cite{lipton2015}. Since RNNs are trained by back-propagation through time, a RNN cell is replaced by a LSTM cell to avoid the vanishing gradient problem \cite{lipton2015}.

Let $x_{t}$ and $h_{t}$ denote the input data and the hidden state at time $t$, respectively. LSTM performance is controlled by three gates:  ($1$) the forget gate $f_t$ selects the information to keep or forget, ($2$) the input gate $i_t$ controls the flow of the input, and ($3$) the output gate $o_t$  that calculates the output of the given updated cell (Figure~\ref{lstmcell}). Formulas governing the operations in the LSTM cell are as follows \cite{lipton2015}:
\begin{equation}
\vspace{-0.02 in}
    f_t = \sigma (W_f . [h_{t-1},x_t]+b_f)
\end{equation}
\begin{equation}
\vspace{-0.02 in}
    i_t = \sigma (W_i . [h_{t-1},x_t]+b_i)
\end{equation}
% \begin{equation}
%     \tilde{C_{t}} = tanh(W_C . [h_{t-1},x_t]+b_C)
% \end{equation}
\begin{equation}
\vspace{-0.02 in}
    C_t = f_t\circ C_{t-1}+i_t\circ \text{tanh}(W_C . [h_{t-1},x_t]+b_C)
\end{equation}
\begin{equation}
\vspace{-0.02 in}
    o_t = \sigma (W_o . [h_{t-1},x_t]+b_o)
\end{equation}
\begin{equation}
\vspace{-0.02 in}
    h_t = o_t\circ \text{tanh}(C_t)
\end{equation}
where $W_{(.)}$, and $b_{(.)}$ are weights and biases, respectively. $C_t$ is the cell state at time $t$. The operator $\circ$ denotes the element-wise vector multiplication.
A bidirectional LSTM (BiLSTM) consists of two LSTM blocks that allow the layer to receive information from the sequential input data simultaneously in forward and backward directions. The output of each layer is the concatenation of outputs of two LSTM blocks, i.e, $h_{i}=[\overrightarrow{h_f},\overleftarrow{h_b}]$ where $\overrightarrow{h_f}$ and $\overleftarrow{h_b}$ correspond to the forward and backward hidden states, respectively \cite{lipton2015}.

% M up to here
 In several sequence-based applications such as semantics analysis, natural language processing, and medical imaging, certain time steps of the input data might contain the most discriminative information, and attention mechanism addresses this issue by focusing on specific time steps \cite{bahdanau2014}. In this mechanism, the most discriminative task-related features are calculated by multiplication of outputs of hidden states by trainable weights. The output of the attention layer, $v$, is computed as bellow:
\begin{equation}
\vspace{-0.01 in}
    v=\sum_{i}{\alpha_{i} h_i}
    \end{equation}
    \begin{equation}
    \vspace{-0.01 in}
    \alpha_{i}= \frac{\exp(W h_i+b)}{\sum_{j}\exp(W h_j+b)}
\end{equation}
where $h_{i}$ denotes the output of the $i\_th$ LSTM layer, and $W$ and $b$ are trainable parameters.
% \subsection{Transfer Learning and Fine Tuning}
% \label{ssec:TL}
% The goal of transfer learning (TL ) is to test our model ability in real-life conditions where the available amount of the labeled data for each similar task is limited. TL helps to improve the learning capability of the target data by using the the knowledge of source domain. In this study, we investigate transferring the learned model parameters with the assumption that individual models across different datasets with similar tasks should share some parameters. Firstly, the model is fully trained using the source dataset. Second, we peruse different layouts to tune the pre-trained network. Figure~\ref{fig3} demonstrates $5$ different schemes that we consider in the calibration (fine-tuning) session. Due to the inspiration by computer vision that in deep neural networks generic features are learned in lower layers and limited labeled target data we simulate a plug \& play device to specify the optimum scheme. 

\subsection{Training, Optimization, and Transfer Learning}
\label{ssec: training}
% Our goal is to show that the proposed SS-STANN architecture not only outperforms the state-of-the-art but also the trained representations could be applied in similar EEG tasks. 
In order to examine the performance of the proposed network, the features of STE and RAN block are fused and fed to the dense layer. Next, the encoded representation is fed to the final softmax classifier. The cross-entropy loss, $\mathcal{L}$, is calculated as follows:
\begin{equation}
\mathcal{L}=\displaystyle\sum_{i} {Y}_i log \hat{Y}_i,
\end{equation}
where $Y_i$ is the ground truth emotion label for each data sample and $\hat{Y}_i$ is the predicted label. Finally, the weights and the biases are trained with batch gradient descent. 

The trained model is used to perform the supervised emotion classification task. Additionally, in the case of medical imaging in general and EEG-based diagnosis in particular, there exist legitimate interests and needs for using such a training model to solve a similar problem with insufficient training data \cite{zhang2021combining, li2021can, wan2021review}. To this end, the model has to be trained on the whole data to get the transferable model parameters. The target data would never be seen in the training phase.

To investigate the possibility of using this trained network in similar EEG-based emotion recognition tasks, we propose and implement a transfer learning (TL) approach. The goal of TL is to test our model ability in real-life conditions where the available amount of labeled data is not sufficient. TL helps to improve the learning capability of the target data by leveraging the knowledge of the source domain. In this study, we investigate transferring the learned model parameters assuming that individual models across different datasets with similar tasks should share some parameters. Firstly, the model is fully trained using sufficient labeled data (source dataset). Second, we peruse different schemes to tune the pre-trained network via the target dataset. The source and target datasets involve EEG-based emotion recognition experiments with different stimuli.

\begin{figure*}[t]
\centering
    \begin{minipage}[c]{.3\textwidth}
    \centering
    \includegraphics[width=0.8\textwidth]{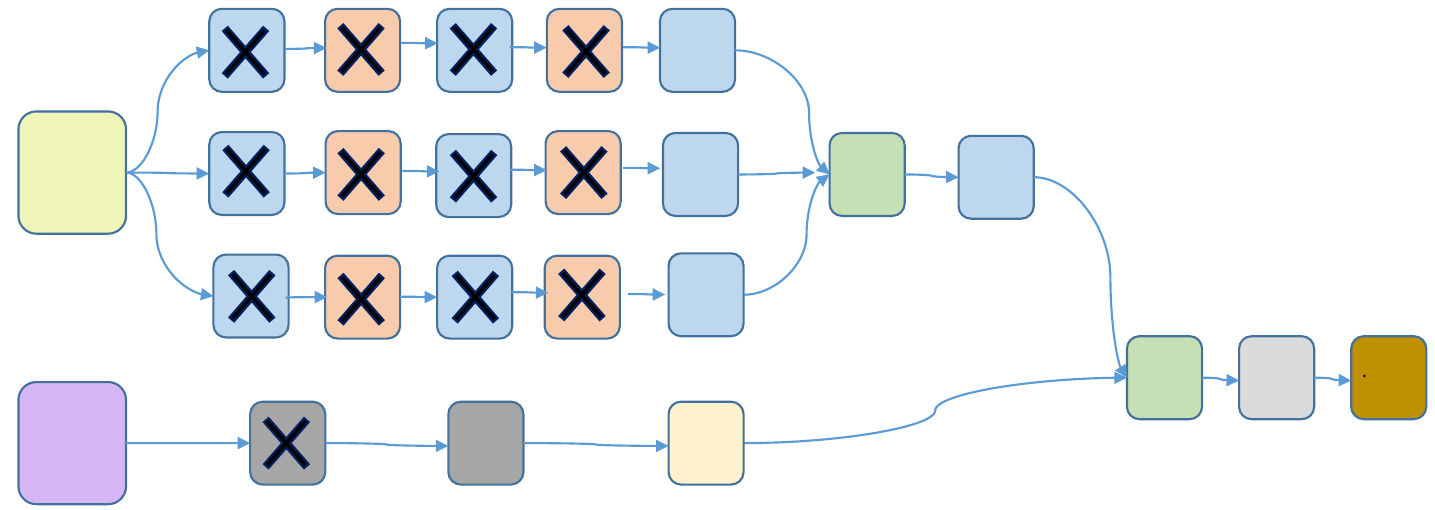}
    \end{minipage}
    \begin{minipage}[c]{.3\textwidth}
    \centering
    \includegraphics[width=0.9\textwidth]{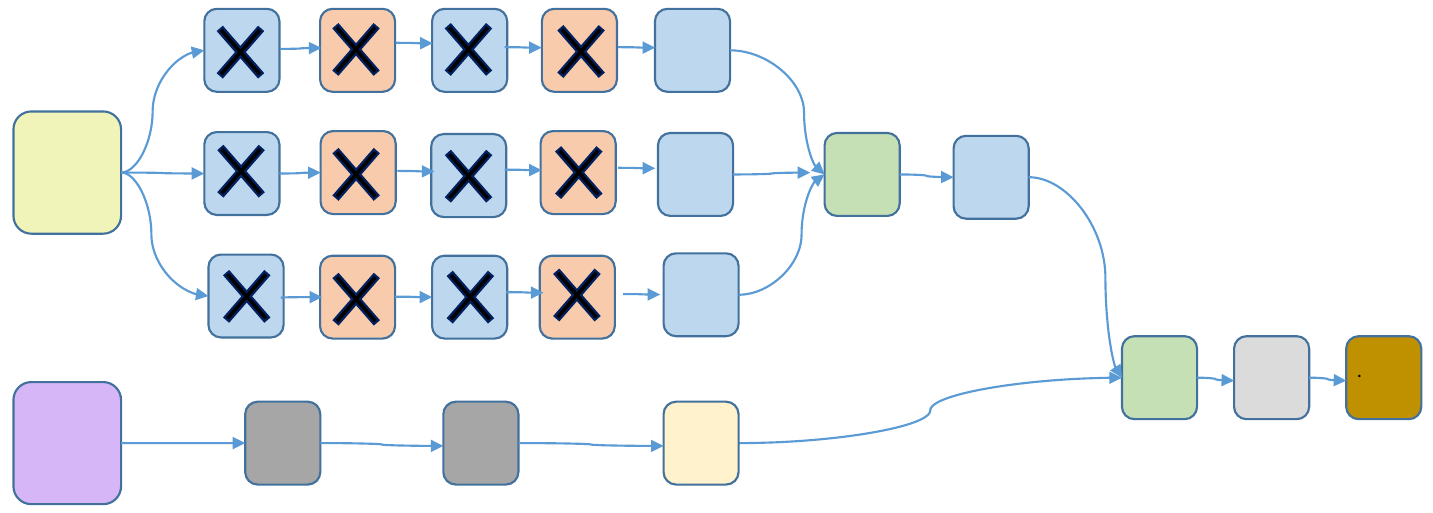}
    \end{minipage}
    \begin{minipage}[c]{.3\textwidth}
    \includegraphics[width=0.9\textwidth]{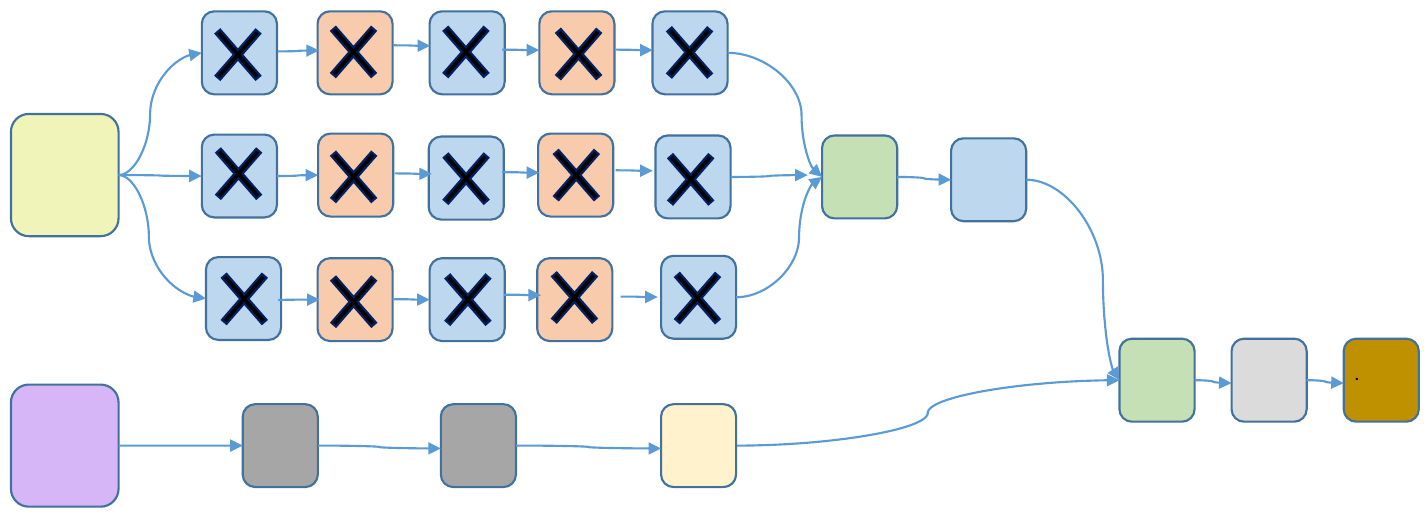}
    \centering
    \end{minipage}
    \begin{minipage}{0.3\textwidth}
 \centering
\text{(a)}
\end{minipage}
\begin{minipage}{0.3\textwidth}
 \centering
\text{(b)}
\end{minipage}
\begin{minipage}{0.3\textwidth}
 \centering
\text{(c)}
\end{minipage}
    \begin{minipage}[c]{.3\textwidth}
    \includegraphics[width=0.9\textwidth]{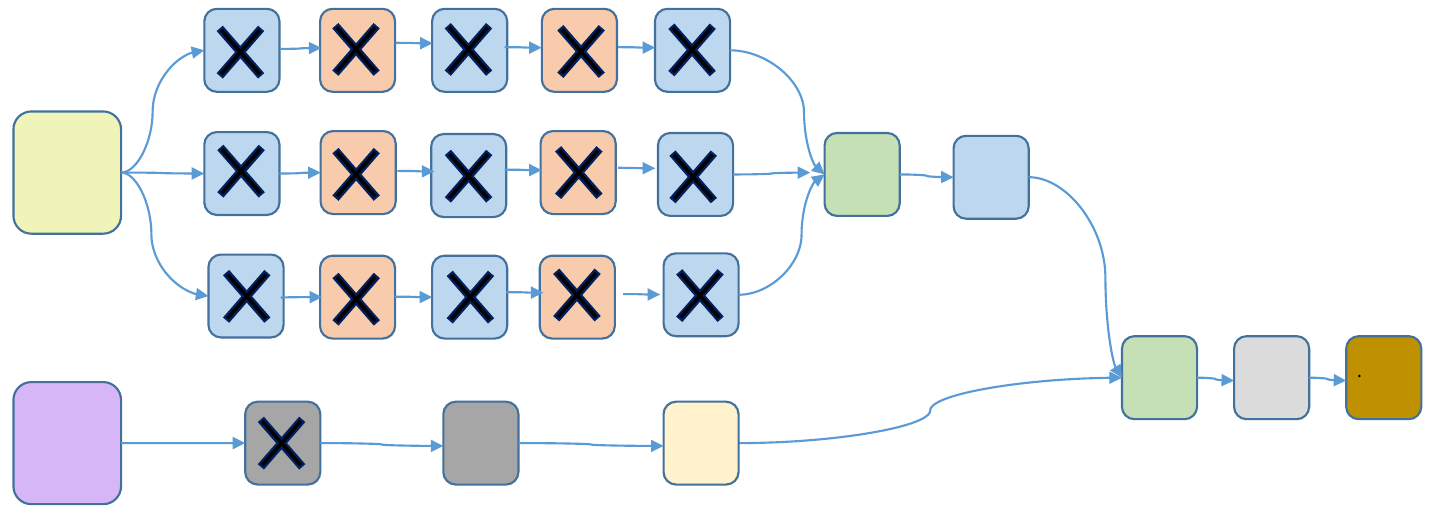}
    \centering
    \end{minipage}
    \begin{minipage}[c]{.3\textwidth}
    \includegraphics[width=0.9\textwidth]{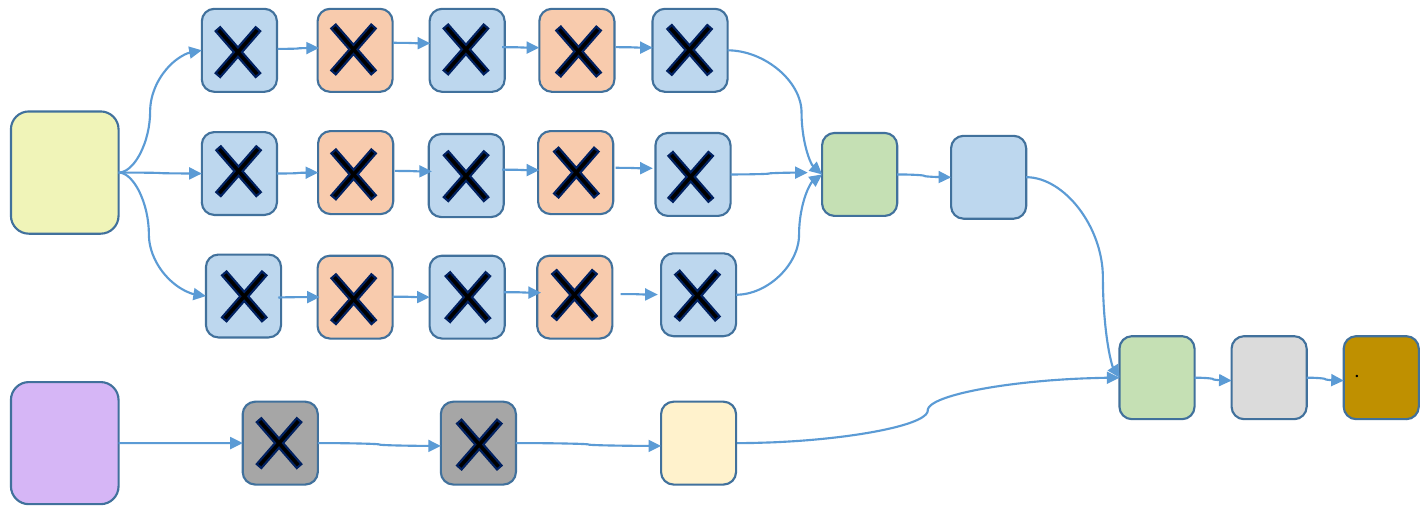}
    \centering
    \end{minipage}
    \begin{minipage}{0.4\textwidth}
 \centering
\text{(d)}
\end{minipage}
\begin{minipage}{0.4\textwidth}
 \centering
\text{(e)}
\end{minipage}
    \caption{The proposed transfer learning schemes. The blocks marked with a black cross are the ones that remain frozen during fine tuning.}
    \label{fig3}
\end{figure*}

Figure~\ref{fig3} demonstrates five different schemes that we consider in the calibration (fine-tuning) session. The TL schemes in our STE blocks are inspired by observations in CNN-based TL frameworks in computer vision where usually the later network layers are retrained, as the earlier layers are responsible for generic features \cite{teuwen2020convolutional}. In our work, we consider different retrainable cells in both STE and RAN blocks. In particular, going from scheme (a) to scheme (e), we change the status of exactly one layer either to retrainable or non-retrainable (frozen) at each step. In each scheme, blocks marked with a cross are left unchanged during fine-tuning of the network. The number of retrainable parameters in scheme (a) to scheme (e) is equal to $239100$, $311420$, $280295$, $207975$, and $53735$, respectively. 

Due to the variations in inter-dataset samples, we use a small part of the new data (target data), $\mathbf{N}$, to calibrate and fine-tune our pre-trained model. Since the amount of calibration data is limited, we scale down the initial learning rate to avoid clobbering in initialization \cite{girshick2015region}. Scaling down the initial learning rate $\eta$ with the scale $\alpha$ can be defined as:
\begin{equation}
    \Phi^{i+1}=(1-\alpha)\Phi^{i}+\alpha(\Phi^{i}-\eta\frac{\partial\mathcal{L}}{\partial\Phi})
\end{equation}
where $\Phi^{i}$ is the trainable parameters at the $i\_th$ iteration and $\mathcal{L}$ is the cross-entropy loss function. Here, $\alpha$ is set to $0.1$.

\section{Dataset and Implementation}
\label{sec:exp}
\subsection{Datasets}
\subsubsection{DEAP Dataset}
\label{sssec:deap}
The DEAP dataset was recorded from $32$ individuals each having rated $40$ music videos for $60$ s \cite{koelstra2011deap}. After each video, the participants performed a self-assessment to show their emotional states by rating the level of valence, arousal, dominance, and liking from $1$ to $9$. The physiological recordings consist of $32$ channels of EEG signals and $8$ channels of peripheral physiological data. Here, we just consider the EEG recordings of each trial. The preprocessing scheme is as follows: 
1) down-sampling the data to $128$ Hz, 2) averaging to the common reference, 3) removing electrooculography (EOG) artifacts, and 4) applying a band-pass filter with the frequency range of $[4, 45]$ Hz. Accordingly, each recording contains $3$ s pre-trial relaxing phase followed by $60$ s trial data.
% EEG data were down-sampled to
% $128$ Hz, averaged to the common reference, electrooculography (EOG) artifacts were eliminated, and a bandpass filter from $4.0$ to $45.0$ Hz was applied. 
% Each EEG recording thus contains $60$ s trial data, in addition to the $3$ s baseline data during which participants were told to relax.
\subsubsection{DREAMER Dataset}
% \label{sssec:target}
% In this work we investigate the performance of the TL on two different target dataset as follows:
% \noindent \textbf{DREAMER.}  
The DREAMER dataset was recorded with a $14$-channel, Emotiv EPOC wireless EEG headset \cite{katsigiannis2017dreamer}. The data were recorded with a $128$ Hz sampling frequency from $23$ participants while watching $18$ film clips. Each film clip lasted $65$ to $393$ s to elicit the emotional states. Before data collection, participants watched a neutral film clip to neutralize the emotional state. Data collected while watching this neutral film served as the baseline. Participants were asked to assess their emotional states by rating valence, arousal, and dominance levels in each video from $1$ to $5$. To have consistency with the DEAP dataset, each recording contains $3$ s pre-trial relaxing phase followed by $60$ s trial data and is band-pass filtered from $4.0$ to $45.0$ Hz.

% $60$ s of trial data, in addition to $3$ s baseline data, were passed through a band-pass filter with bandwidth from $4.0$ to $45.0$ Hz.  

\begin{table*}[t]
\centering
\caption{Details of the implemented STE block.}
\label{table:2}
\footnotesize{
\begin{tabular}[t]{ c c c c c c|| c c c c c c}
\hline
\textbf{Block} & \textbf{Type} & \textbf{Filters} &  \textbf{Kernel} & \textbf{Output Shape}&
\textbf{Parameters}&\textbf{Block} & \textbf{Type} & \textbf{Filters} &  \textbf{Kernel} & \textbf{Output Shape}&
\textbf{Parameters}\\
\hline
IN1  & CNN Input&$-$&$-$&$32\times{}128\times{}1$&$-$& C2\_3&Conv2D+BN&$80$&$1\times{}1$&$16\times{}64\times{}80$&$3600$\\
\hline
C1\_1  & Conv2D+BN&$25$&$5\times{}5$&$32\times{}128\times{}25$&$750$& P2\_1&Pool2D&$-$&$2\times{}2$&$8\times{}32\times{}50$&$-$\\
\hline
C1\_2  & Conv2D+BN&$30$&$5\times{}5$&$32\times{}128\times{}30$&$900$& P2\_2&Pool2D&$-$&$2\times{}2$&$8\times{}32\times{}60$&$-$\\
\hline
C1\_3  & Conv2D+BN&$40$&$3\times{}3$&$32\times{}128\times{}40$&$560$& P2\_3&Pool2D&$-$&$2\times{}2$&$8\times{}32\times{}80$&$-$\\
\hline
P1\_1  &Pool2D&$-$&$2\times{}2$&$16\times{}64\times{}25$&$-$& C3\_1&Conv2D+BN&$25$&$3\times{}3$&$8\times{}32\times{}25$&$11375$\\
\hline
P1\_2  &Pool2D&$-$&$2\times{}2$&$16\times{}64\times{}30$&$-$& C3\_2&Conv2D+BN&$30$&$3\times{}3$&$8\times{}32\times{}30$&$16350$\\
\hline
P1\_3  &Pool2D&$-$&$2\times{}2$&$16\times{}64\times{}40$&$-$& C3\_3&Conv2D+BN&$40$&$3\times{}3$&$8\times{}32\times{}40$&$3400$\\
\hline
C2\_1  & Conv2D+BN&$50$&$3\times{}3$&$16\times{}64\times{}50$&$11500$& Con1&concatenate&$-$&$-$&$8\times{}32\times{}95$&$-$\\
\hline
C2\_2  & Conv2D+BN&$60$&$3\times{}3$&$16\times{}64\times{}60$&$16500$& C4&Conv2D&$1$&$1\times{}1$&$8\times{}32\times{}1$&$96$\\
\hline
\end{tabular}
}
\label{table1}
\end{table*}
% \noindent \textbf{Emotion Word Dataset.}
\subsubsection{Emotional English Word Dataset (EEWD)}
 Data collection was performed via $64$ Ag/AgCl active electrodes located over the scalp based on the $10$-$10$ International Electrode Placement System while participants were rating emotional words. Participants provided signed informed consents in accordance with the Sabanci University Research Ethics Council guidelines. EEG data were recorded using BioSemi ActiveTwo systems (Biosemi Inc., Amsterdam,
the Netherlands) in a dimly lit EEG room within a Faraday cage. A dataset of highly-arousing English words was formed such that $65$ negative words (arousal$>6$ and valence$<3$) and $63$ positive words (arousal$>6$ and valence$>7$) were selected from the Affective Norms for English Words (ANEW) dataset \cite{bradley2010affective}. Details of compiling this small dataset of Original English List (OEL) is presented in more details in \cite{torkamani2019emotionality}. Thirty native Turkish speakers participated in the experiment where English served as their secondary language. The experiment consisted of four blocks and each block contained $32$ randomly selected words. Each word was presented for $1$ s and then the participants were asked to rate the valence and arousal in the range of $1$ to $9$ using a set of pictorial self-assessment manikins (SAM) \cite{torkamani2019emotionality}. The data of two participants were discarded due to technical problems. The recorded signals were down-sampled from $2048$ Hz to
$128$ Hz, EOG artifacts were removed via independent component analysis (ICA), and a bandpass filter from $4.0$ to $45.0$ Hz was applied. All preprocessing steps were conducted using the EEGLAB toolbox \cite{delorme2004eeglab}.

\subsection{Implementation of the Proposed Network}
\label{ssec:propos}
The proposed network consists of two parts that operate in parallel \cite{shadiembc2021}. The input of STE, IN$1$, has the dimensions of n$\times$k$\times$1 where $k$ is set to $128$ corresponding to the $1$ s slicing window length. In order to select the STE parameters, we perform the grid search on the subset of the kernel sizes, $[9,7,5,3]$, and the parameters corresponding to the best performance on the training data samples selected. Table \ref{table1} provides the details of the STE structure. Each average-pooling layer is followed by a dropout layer. The dropout probability rates for each dropout layer are set to $0.5$ and $0.4$, respectively. In order to prevent edge information loss, the same zero-padding technique is used in each convolution (conv) operation. Here, we adopt rectified linear unit (ReLU) which has been used in many related CNN-based applications. After concatenating the outputs of all columns, a $1\times{}1$ conv filter is applied to compute the spatial feature maps. 

The input of the RAN block, IN$2$, has the size of $k\times n$. RAN consists of two BiLSTM layers with the same hidden layer size. The hidden layer and time steps are set to $80$ and $128$, respectively. We choose hyperbolic tangent (\text{tanh}) as the activation function for all BiLSTM cells. Each pair of forward and backward LSTM cells is followed by a dropout layer with probability rates of $0.3$ and $0.2$, respectively. BiLSTM outputs are then used as the input of the attention mechanism. 

The spatial and temporal representations extracted from STE and RAN blocks are flattened and concatenated. Next, we apply a fully connected layer where the number of hidden units is $128$. In the end, we exploit the SoftMax operation to obtain classification labels.

\section{Experiments}
\label{ssec:class}
% In this subsection, we evaluate the performance of the proposed architecture. First, we discuss the performance of SS-STANN on the DEAP dataset which also serves as the source dataset. Second, we evaluate TL results on the target dataset.

\subsection{Results and Analysis on the DEAP Dataset}
\label{ssec:deapresults}
 In this section, we investigate the performance of the proposed DL architecture on the DEAP dataset. The DEAP dataset involves subjects watching long, continuous videos, hence trials can be defined arbitrarily as smaller segments from this dataset. While the DEAP dataset defines $60$ s EEG intervals as trials, various methods have used different intervals as samples for training and testing. In this work, the trial data are baseline corrected and a non-overlapping sliding window with a length of $1$ s is applied to slice the $60$ s trials. The final size of each data sample equals $32\times{}128$ where $32$ is the number of the EEG nodes and $128$ is the number of time samples. Thus, the data for each participant consists of $40\times{}60= 2400$ data samples. While our experiments and those of other methods we compare against have allowed non-overlapping $1$ s data samples from any $60$ s trial to be assigned to training or test sets randomly, a different approach could be to assign all $1$ s samples from a particular trial to the training or test set. In order to explore different frequency bandwidths, each data sample is filtered into five subbands: theta, alpha, beta, gamma, and wide-band. In order to validate the performance of the proposed method, we consider three binary classification problems, i.e., high-versus-low valence, high-versus-low arousal, and high-versus-low dominance. Considering a threshold of $5$, we quantize the $9$-level ratings of valence, arousal, and dominance into two levels to obtain a binary problem. The model is trained using subject-dependent $10$-fold cross-validation (CV). The validation process is repeated $10$ times and the average classification performance is reported. Adam optimizer \cite{kingma2014adam} is used to minimize the cross-entropy between the predicted labels and true labels. The epochs and batch-size are picked as $50$ and $300$, respectively. Grid search is applied to select the hyper-parameters that maximize the average classification accuracy based on training data.  

\subsubsection{Ablation Study}
\label{sssec:deapablation}
To analyze the impacts of using the station-temporal encoding, recurrent network, and attention mechanism, we have performed an ablation study by considering each block of the proposed STANN, namely STE and RAN as baseline models and evaluating their performance separately using the aforementioned parameters. In order to assess the effect of graph smoothing, we calculate the graph-smoothed signals and assess the network performance in frequency subbands and present the performance results of the proposed method with two different input modalities, i.e., raw EEG data and smooth EEG data. Since the GFT computing and smoothness are not dependent on a specific task or subject, this would not interfere with the automated operation of our model. Details of the ablation experiments are provided in Table~\ref{ablationexperiments}.

Tables \ref{table2}, \ref{table3}, and \ref{table4} show the obtained average accuracies and standard deviations (SD) for binary valence, binary arousal, and binary dominance classification for the baseline models and the proposed method based on data from different frequency subbands with different inputs. To address cases involving unbalanced data, we also report F$1$-scores for the proposed model in all three classification problems. SS$2$-(.) and SS$4$-(.) correspond to graph-smoothed signals with $2$-NN and $4$-NN adjacency matrices, respectively. 
\begin{table}[t]
\centering
\caption{Ablation experiments.}
\footnotesize{
\begin{tabular}[t]{ l|c c }
\hline
\textbf{Method} & \multicolumn{1}{c}{Network Components}& \multicolumn{1}{c}{Input}\\
\hline
RAN  & recurrent attention
network & raw data\\
SS2-RAN  & recurrent attention
network & smoothed data with 2-NN\\
STE  & spatio-temporal encoding & raw data\\
SS2-STE  & spatio-temporal encoding & smoothed data with 2-NN\\
STANN  & RAN+STE & raw data\\
SS2-STANN  & RAN+STE & smoothed data with 2-NN\\
SS4-STANN  & RAN+STE & smoothed data with 4-NN\\
\hline

\hline
\end{tabular}
}
\label{ablationexperiments}
\end{table}
\begin{table}[t]
\centering
\caption{Ablation study on binary valence classification accuracies ($\%$) on the DEAP dataset (accuracy/standard deviation). For SS2-STANN and SS4-STANN, the F1-score is shown in blue.}
\footnotesize{
\begin{tabular}[t]{ l|c c c c c }
\hline
\textbf{Method} & \multicolumn{1}{c}{$\theta$} & \multicolumn{1}{c}{$\alpha$} &  \multicolumn{1}{c}{$\beta$} & \multicolumn{1}{c}{$\gamma$}&
\multicolumn{1}{c}{wide-band}\\
\hline
RAN  &$84.6/5.5$&$85.5/4.4$&$87.0/5.6$&$83.5/7.1$&$89.0/5.5$\\
\hline
SS2-RAN &$85.2/4.7$&$85.9/4.3$&$88.4/5.1$&$84.2/6.6$&$90.1/5.0$\\
\hline
STE &$86.1/4.5$&$86.0/3.6$&$91.4/2.5$&$83.2/5.0$&$93.3/2.2$ \\
\hline
SS2-STE &$87.8/3.8$&$89.2/5.2$&$91.6/4.0$&$85.2/5.8$&$93.8/1.9$ \\
\hline
STANN &$88.1/3.6$&$88.5/3.1$&$91.2/2.2$&$86.6/4.5$&$94.4/2.1$\\
\hline
SS2-&$\bf{89.1}$/$4.7$&$91.2/3.7$&$94.9/2.4$&$91.1/3.7$&$94.7/2.4$\\STANN &$\color{blue}0.89$&$\color{blue}0.89$&$\color{blue}0.93$&$\color{blue}0.89$&$\color{blue}0.94$\\
\hline
SS4-&$88.9/4.7$&$\bf{91.7}$/$3.9$&$\bf{95.1}$/$2.2$&$\bf{91.3}$/$3.8$&$\bf{95.6}$/$1.9$\\STANN &$\color{blue}0.89$&$\color{blue}0.89$&$\color{blue}0.94$&$\color{blue}0.89$&$\color{blue}0.94$\\
\hline
\end{tabular}
}
\label{table2}
\end{table}
\begin{table}[t]
\centering
\caption{Ablation study on binary arousal classification accuracies ($\%$) on the DEAP dataset (accuracy/standard deviation).For SS2-STANN and SS4-STANN F1 score is shown in blue.}
\footnotesize{
\begin{tabular}[t]{ l|c c c c c }
\hline
\textbf{Method} & \multicolumn{1}{c}{$\theta$} & \multicolumn{1}{c}{$\alpha$} &  \multicolumn{1}{c}{$\beta$} & \multicolumn{1}{c}{$\gamma$}&
\multicolumn{1}{c}{wide-band}\\
\hline
RAN  &$87.0/4.5$&$87.2/4.6$&$88.1/3.5$&$85.0/3.3$&$90.3/3.6$\\
\hline
SS2-RAN &$87.6/4.1$&$88.2/4.5$&$89.5/3.5$&$85.9/3.2$&$91.8/3.3$\\
\hline
STE &$88.5/4.6$&$87.0/4.0$&$91.6/3.8$&$84.2/3.5$&$93.9/3.8$\\
\hline
SS2-STE &$89.2/4.3$&$88.1/3.6$&$91.9/3.6$&$85.2/3.0$&$94.5/3.1$\\
\hline
STANN &$90.2/4.7$&$89.7/3.7$&$92.5/3.8$&$86.7/3.6$&$94.9/2.8$\\
\hline
SS2-&$92.5/3.2$&$92.6/3.7$&$\bf{95.4}$/$2.3$&$91.5/3.8$&$95.8/1.8$\\STANN &$\color{blue}0.90$&$\color{blue}0.90$&$\color{blue}0.94$&$\color{blue}0.88$&$\color{blue}0.95$\\
\hline
SS4-&$\bf{92.7}$/$4.0$&$\bf{92.6}$/$3.7$&$95.2/2.1$&$\bf{91.6}$/$3.7$&$\bf{97.0}$/$1.7$\\STANN &$\color{blue}0.90$&$\color{blue}0.90$&$\color{blue}0.93$&$\color{blue}0.89$&$\color{blue}0.95$\\
\hline
\end{tabular}
}
\label{table3}
\end{table}
\begin{table}[!t]
\centering
\caption{Ablation study on dominance classification accuracies ($\%$) on the DEAP dataset (accuracy/standard deviation). For SS2-STANN and SS4-STANN, the F1-score is shown in blue.}
\footnotesize{
\begin{tabular}[t]{ l|c c c c c}
\hline
\textbf{Method} & \multicolumn{1}{c}{$\theta$} & \multicolumn{1}{c}{$\alpha$} &  \multicolumn{1}{c}{$\beta$} & \multicolumn{1}{c}{$\gamma$}&
\multicolumn{1}{c}{wide-band}\\
\hline
RAN  &$87.9/5.4$&$88.9/5.1$&$89.0/5.9$&$84.3/7.9$&$89.4/5.4$\\
\hline
SS2-RAN  &$89.0/5.3$&$89.5/4.7$&$89.8/5.5$&$85.6/7.2$&$90.2/4.7$\\
\hline
STE &$91.6/4.5$&$92.5/4.4$&$95.3/2.8$&$88.8/5.7$&$95.5/2.7$\\
\hline
SS2- &$91.9/4.5$&$92.5/4.1$&$95.5/2.6$&$89.3/5.2$&$95.7/2.3$\\
\hline
STANN &$92.2/3.9$&$93.9/3.9$&$96.3/2.8$&$90.4/5.5$&$96.3/2.3$\\
\hline
SS2-&$92.8/3.8$&$93.0/4.2$&$95.5/2.7$&$91.6/4.6$&$96.1/2.4$\\STANN &$\color{blue}0.90$&$\color{blue}0.90$&$\color{blue}0.94$&$\color{blue}0.88$&$\color{blue}0.94$\\
\hline
SS4-&$\bf{92.7}/4.0$&$\bf{93.2}/4.0$&$\bf{95.6}/2.3$&$\bf{91.7}/4.4$&$\bf{96.8}/1.9$\\STANN &$\color{blue}0.89$&$\color{blue}0.90$&$\color{blue}0.94$&$\color{blue}0.88$&$\color{blue}0.95$\\
\hline
\end{tabular}
}
\label{table4}
\end{table}
Results of DEAP dataset classification in Tables~\ref{table2} to~\ref{table4} demonstrate that graph smoothing leads to better performance than the use of raw EEG input data. Moreover, in the majority of classification scenarios, the best performance is driven by wide-band data and SS$4$-STANN. The average classification accuracies for binary valence, arousal, and dominance classification problems based on SS$4$-STANN are $95.6\%$, $97.0\%$, and $96.8\%$, respectively. These results indicate that our proposed method exceeds the baseline models and that graph smoothing enhances the overall classification performance. Moreover, our findings show that beta and wide-band frequency subbands outperform other spectral features in binary classifications problems of high-versus-low valence, arousal, and dominance dimensions which is in line with the role of different frequency subbands in characterizing affective states \cite{ray1985eeg, zheng2015investigating}.

\subsubsection{Comparison with the-state-of-the-art}
We compare the performance of our proposed SS4-SSTANN method for the classification of valence and arousal from wide-band EEG of the DEAP dataset with a number of state-of-the-art solutions namely DCCA \cite{liu2021comparing}, ECLGCNN \cite{yin2021eeg}, DGCNN \cite{song2018eeg}, CVCNN \cite{chen2019accurate}, CRAM \cite{zhang2019convolutional}, ACRNN \cite{tao2020eeg}, S-EEGNet \cite{huang2020s}, and Casc-CNN-LSTM \cite{chen2020emotion}. In DCCA, firstly, DE features are computed in four different frequency bands and then the network computes the representations of two modalities by passing them through multiple stacked layers of nonlinear transformations.ECLGCNN uses the infusion of graph convolutional neural networks with LSTMs while DE of the windowed EEG data are considered as the input. DGCNN\footnote[1]{https://github.com/xueyunlong12589/DGCNN} computes DE features and applies a $2400$ feature vector as the input of the graph CNN. CVCNN utilizes raw EEG and normalized EEG data in combination with PSD features. CRAM extracts spatio-temporal information along with attentive temporal dynamics in a cascaded format. The model uses a CNN layer with a fixed kernel and filter size and the temporal information is extracted from the extracted spatial features which makes the temporal information dependent on the spatial features. ACRNN uses an attention technique to get different weights for each EEG channel followed by a CNN to extract spatial features. S-EEGNet applies the Hilbert–Huang transform to preprocess the EEG data before feeding the data to a separable CNN. Casc-CNN-LSTM applies hybrid convolutional recurrent neural networks by using transformed 1D EEG vector sequences into 2D mesh-like matrices. Table \ref{table5} presents a comparison of the proposed SS4-STANN on wide-band data with the above-mentioned methods from recent literature and demonstrates the superiority of our proposed approach. The reported results are all subject-dependent with a $10$-fold CV except \cite{huang2020s} where the authors performed a $4$-fold CV. 
\begin{table}[t]
    \centering
    \caption{Comparison of the proposed SS4-STANN method with recent state-of-the-art solution for valence and arousal classification from wide-band data of the DEAP Dataset.}
    % \vspace{-0.18in}
    
    \footnotesize{
    \begin{tabular}[t]{c| c  c}
    \hline
\multicolumn{1}{c|}{\textbf{Method}} & \multicolumn{1}{c}{Valence ($\%$)} & \multicolumn{1}{c}{Arousal ($\%$)} \\
        \hline
    {\bf Proposed method with SS4-STANN}               & $\bf{95.6}$ & $\bf{97.0}$\\
    \hline
    DCCA \cite{liu2021comparing} & $84.3$&$85.6$\\
    \hline
    ECLGCNN \cite{yin2021eeg} & $90.5$&$90.6$\\
    \hline
    DGCNN \cite{song2018eeg} & $92.5$&$93.5$\\
    \hline
    CVCNN \cite{chen2019accurate}& $88.8$&$86.9$\\
    \hline
    CRAM \cite{zhang2019convolutional}& $85.5$&$83.6$\\
    \hline
   ACRNN \cite{tao2020eeg}& $89.9$&$88.3$\\
    \hline
    S-EEGNet \cite{huang2020s}& $89.9$&$88.3$\\
    \hline
    Casc-CNN-LSTM \cite{chen2020emotion}& $93.6$&$93.2$\\
    \hline
    \end{tabular}
    % \vspace{-0.05in}
    }
    % \vspace{-0.15in}
    \label{table5}
\end{table}

\begin{figure}[t]
    \centering
    \begin{minipage}{0.05\textwidth}
\text{Col. $1$}
\end{minipage}
    \begin{minipage}[c]{.13\textwidth}
     \centering{\small High valence}
    \includegraphics[width=1\textwidth]{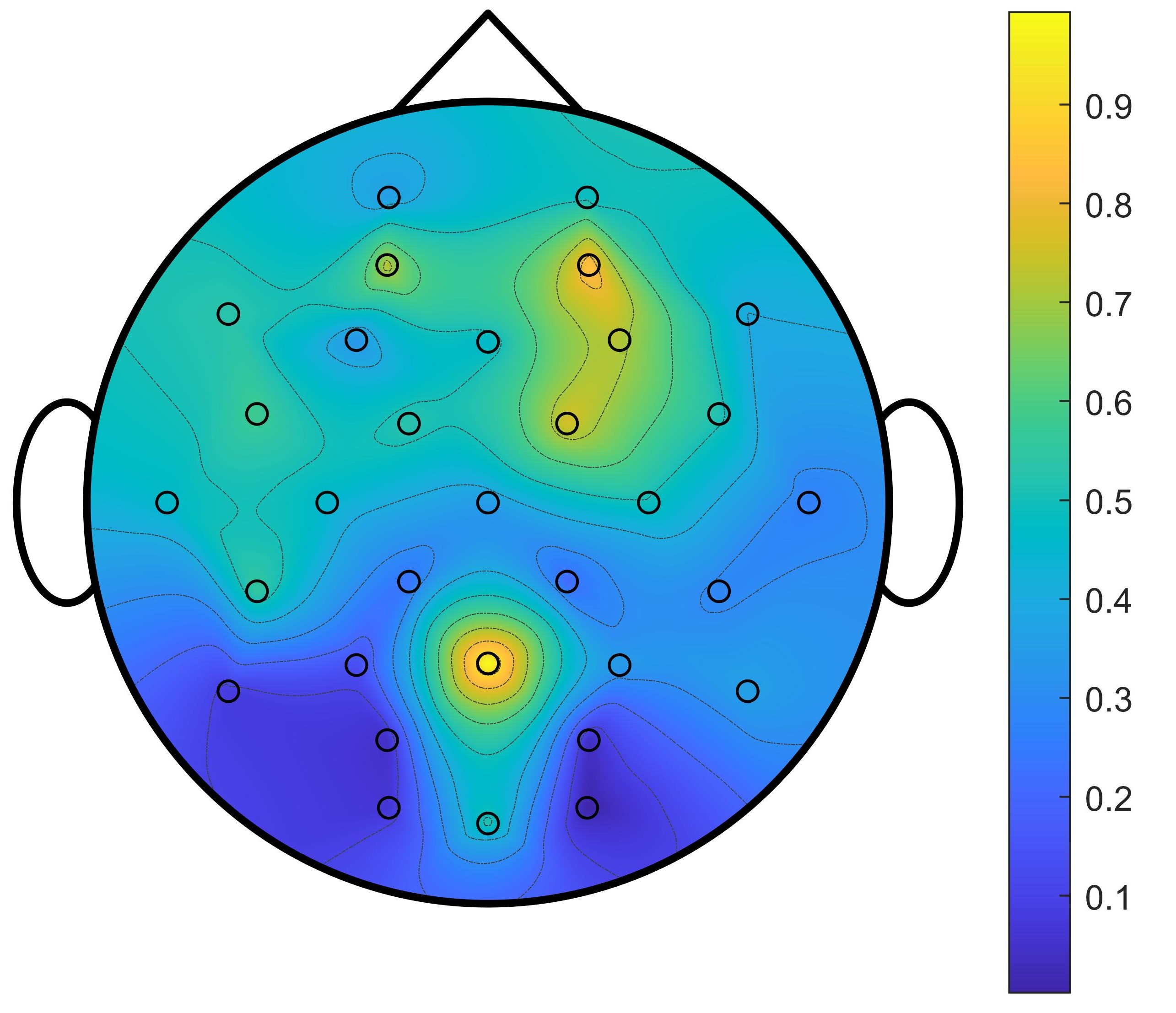}
    \end{minipage}
    \begin{minipage}[c]{.13\textwidth}
    \centering{\small Low valence}
    \includegraphics[width=1\textwidth]{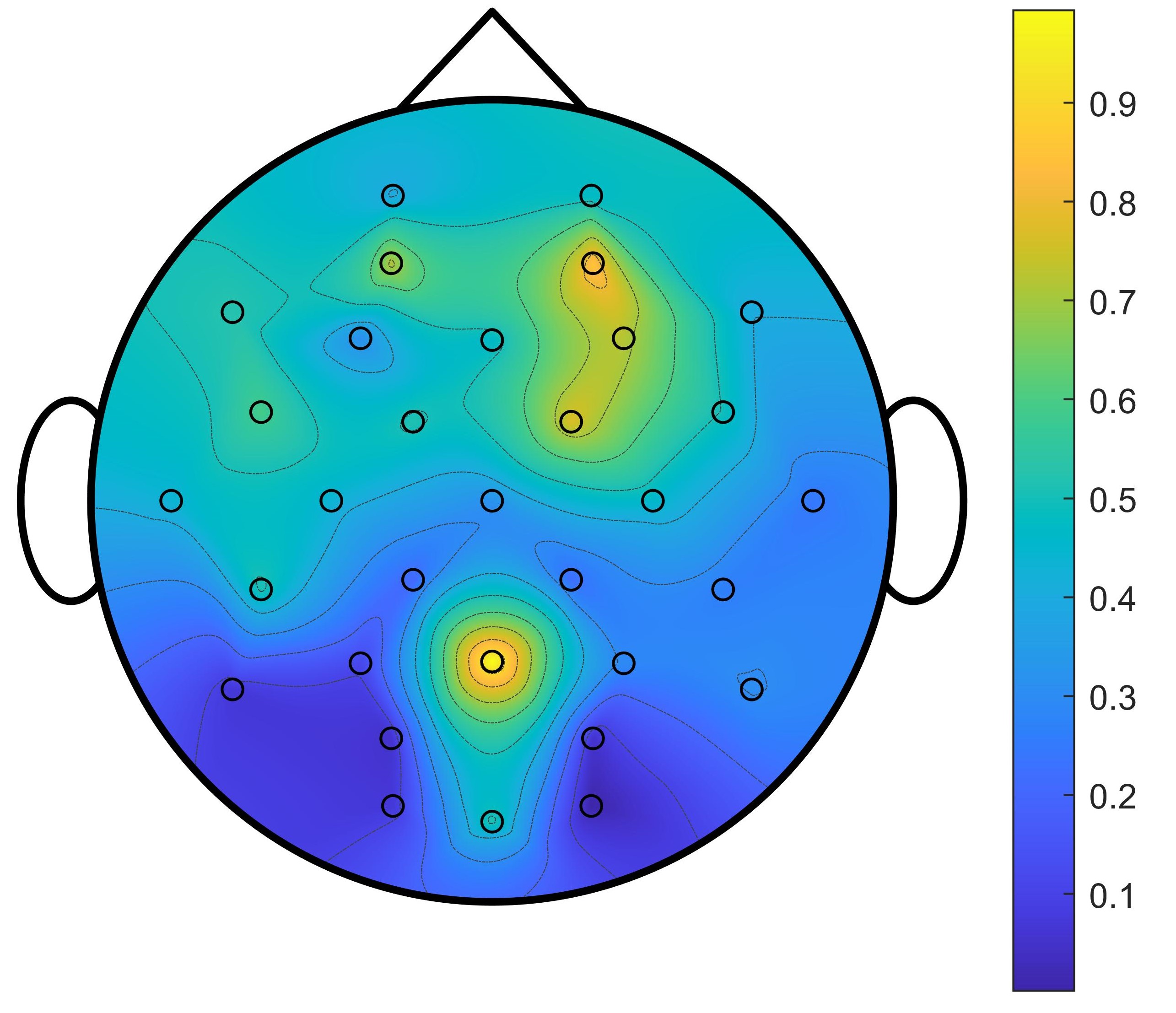}
    \end{minipage}
    \begin{minipage}[c]{.13\textwidth}
    \centering{\small (High-Low) valence}
    \includegraphics[width=1\textwidth]{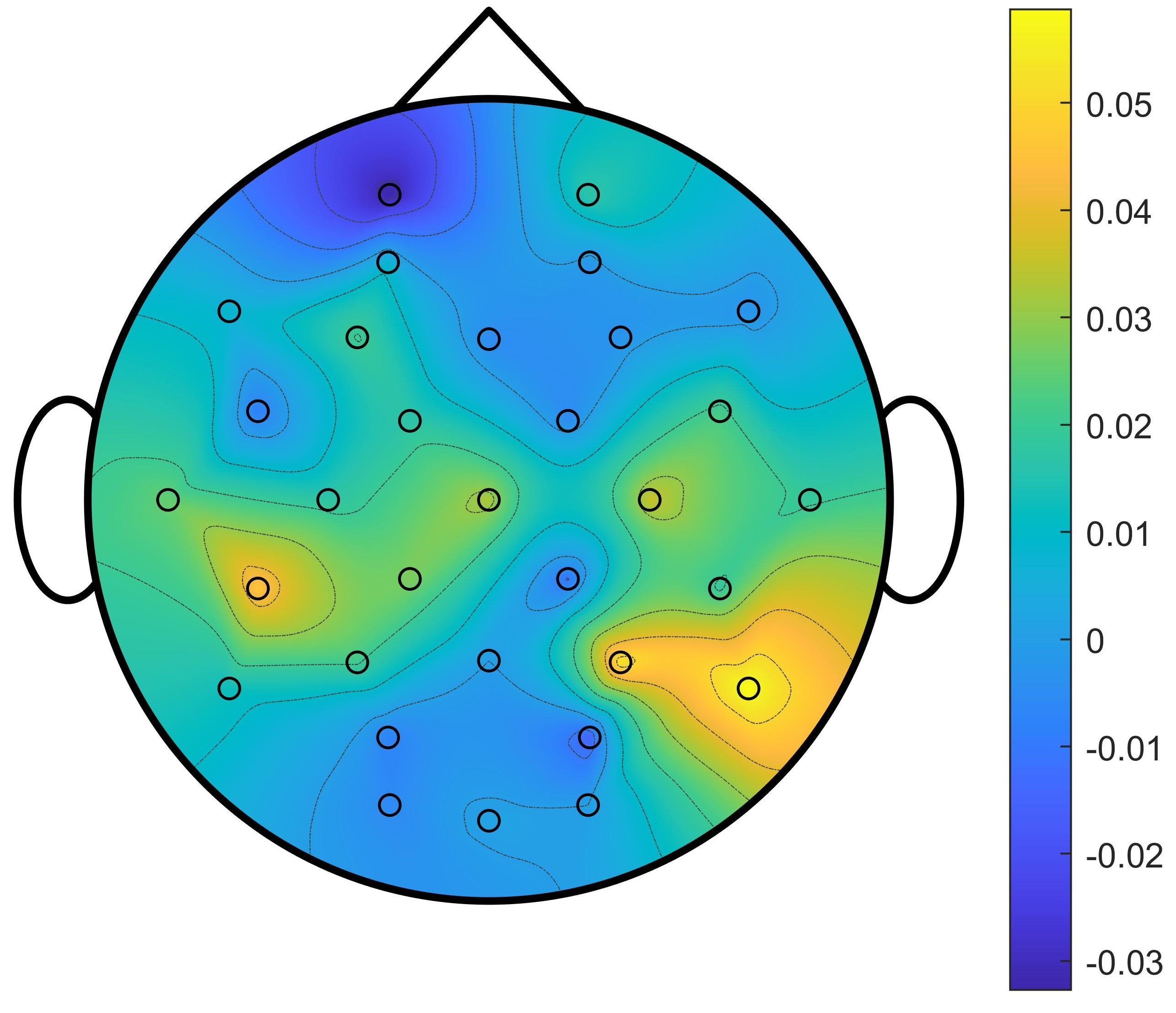}
    \end{minipage}
    \begin{minipage}{0.05\textwidth}
\text{Col. $2$}
\end{minipage}
    \begin{minipage}[c]{.13\textwidth}
    % \centering{\small High arousal}
    \includegraphics[width=1\textwidth]{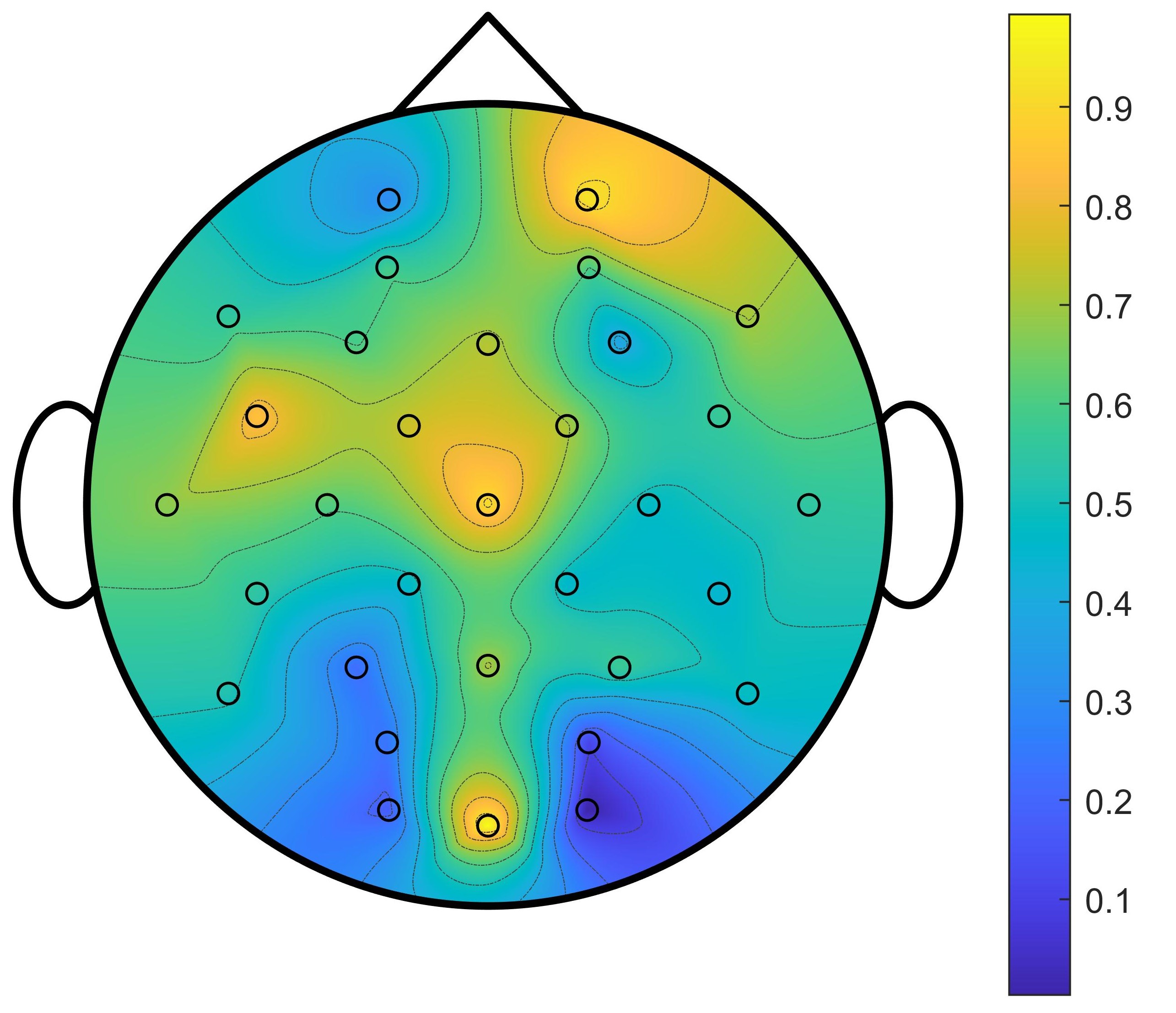}
    \end{minipage}
    \begin{minipage}[c]{.13\textwidth}
    % \centering{\small Low arousal}
    \includegraphics[width=1\textwidth]{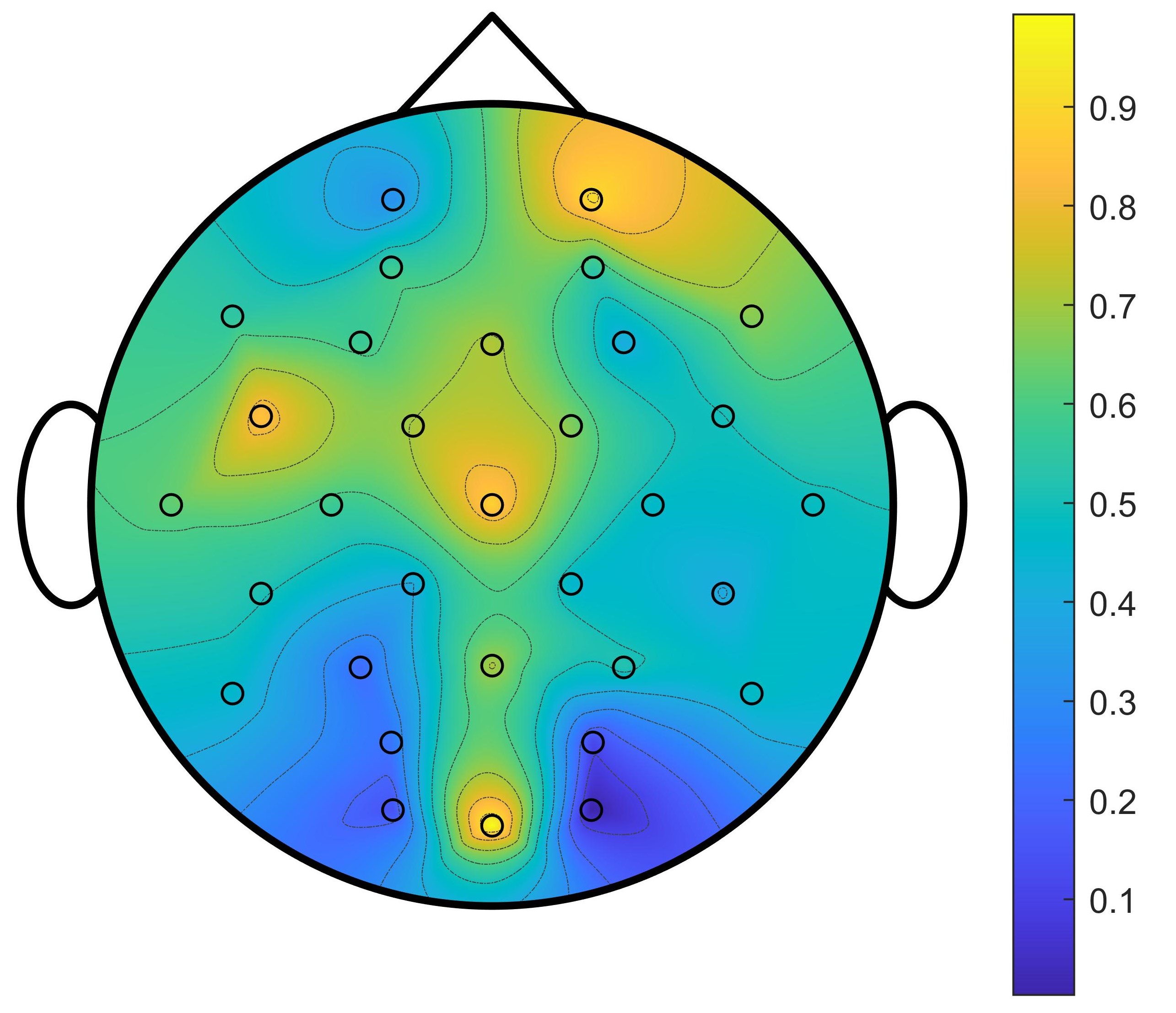}
    \end{minipage}
    \begin{minipage}[c]{.13\textwidth}
    % \centering{\small (High-Low) arousal}
    \includegraphics[width=1\textwidth]{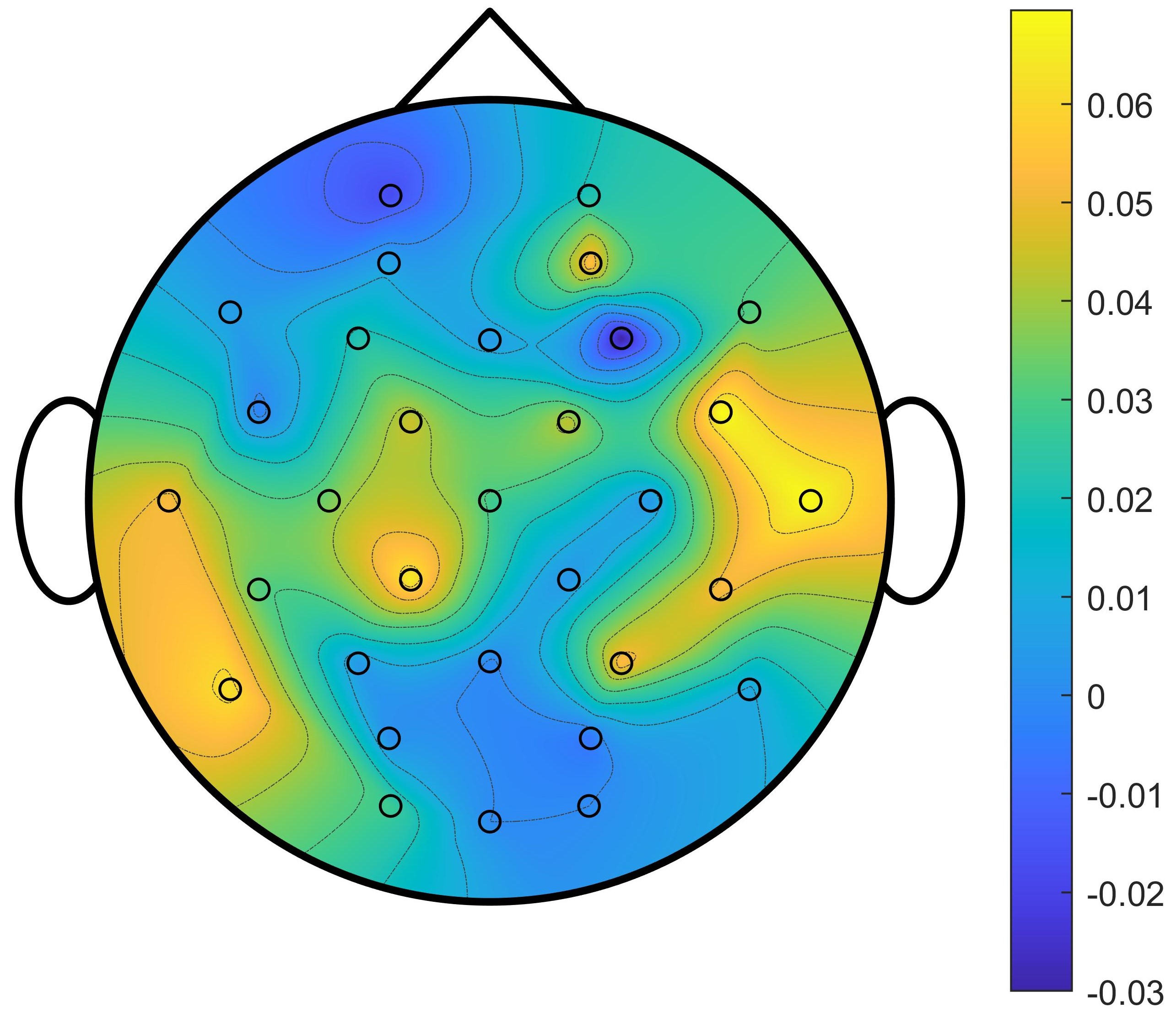}
    \end{minipage}
    \begin{minipage}{0.05\textwidth}
\text{Col. $3$}
\end{minipage}
    \begin{minipage}[c]{.13\textwidth}
    %  \centering{\small High valence}
    \includegraphics[width=1\textwidth]{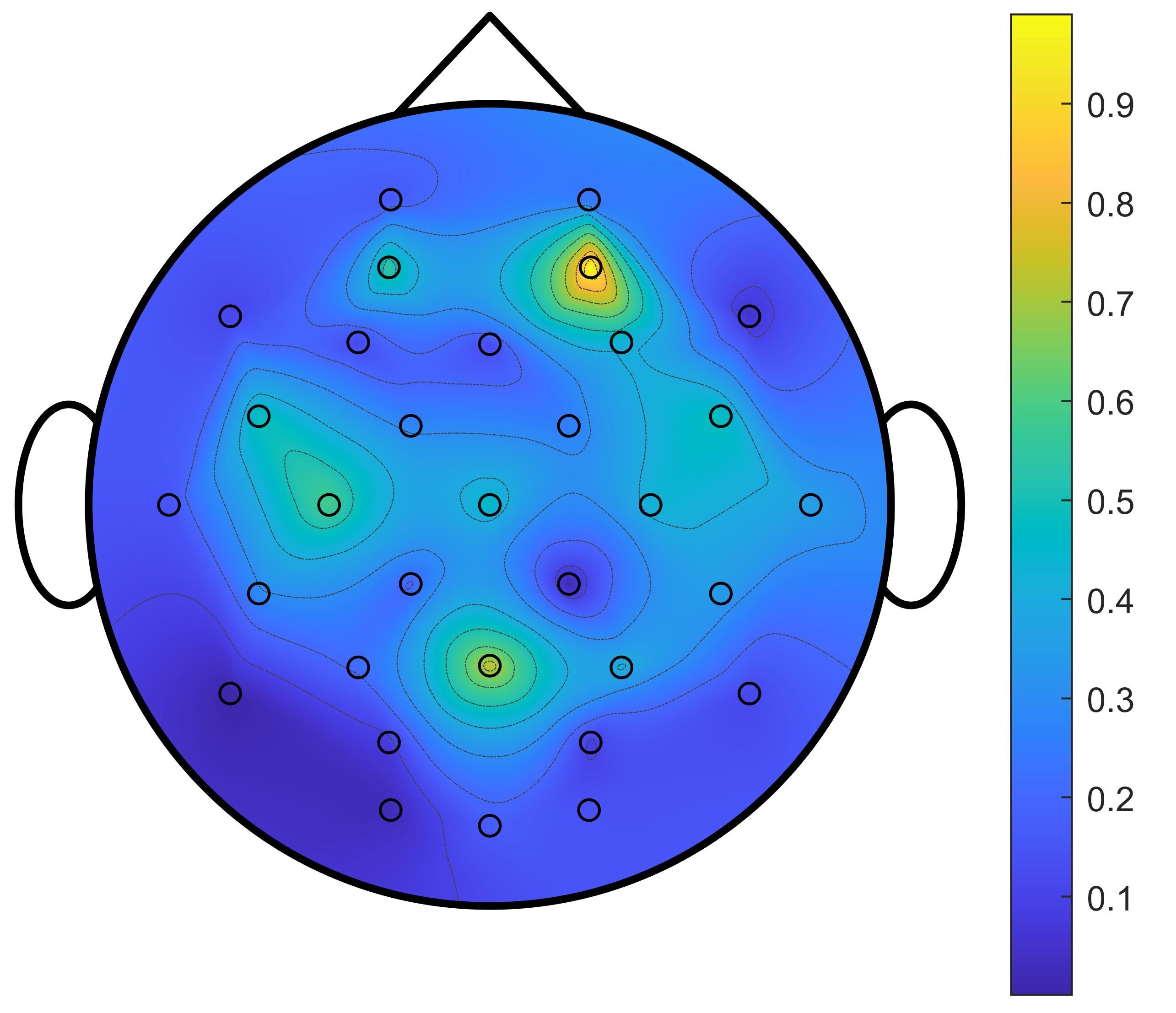}
    \end{minipage}
    \begin{minipage}[c]{.13\textwidth}
    % \centering{\small Low valence}
    \includegraphics[width=1\textwidth]{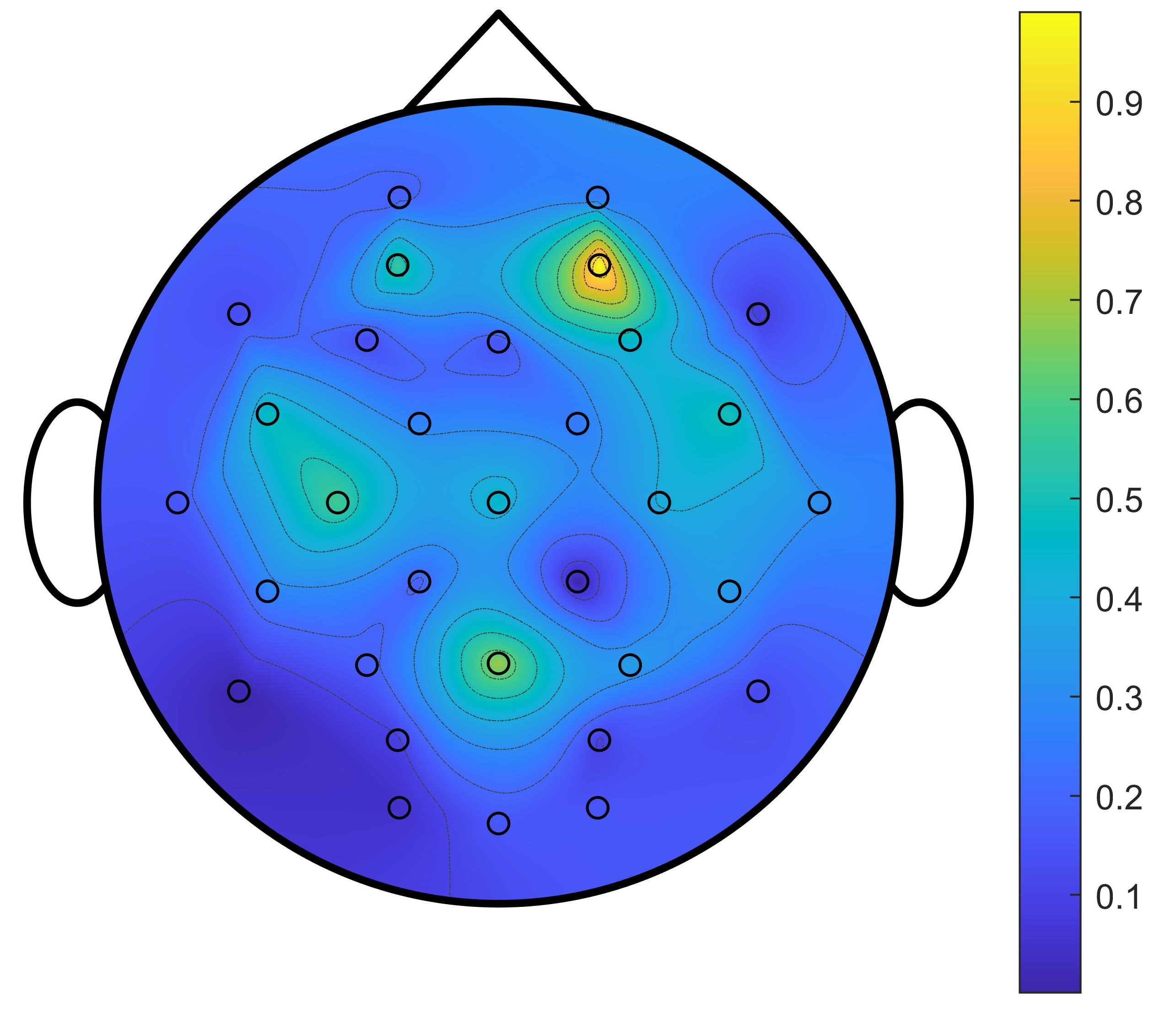}
    % \subcaption{a}
    \end{minipage}
    \begin{minipage}[c]{.13\textwidth}
    % \centering{\small (High-Low) valence}
    \includegraphics[width=1\textwidth]{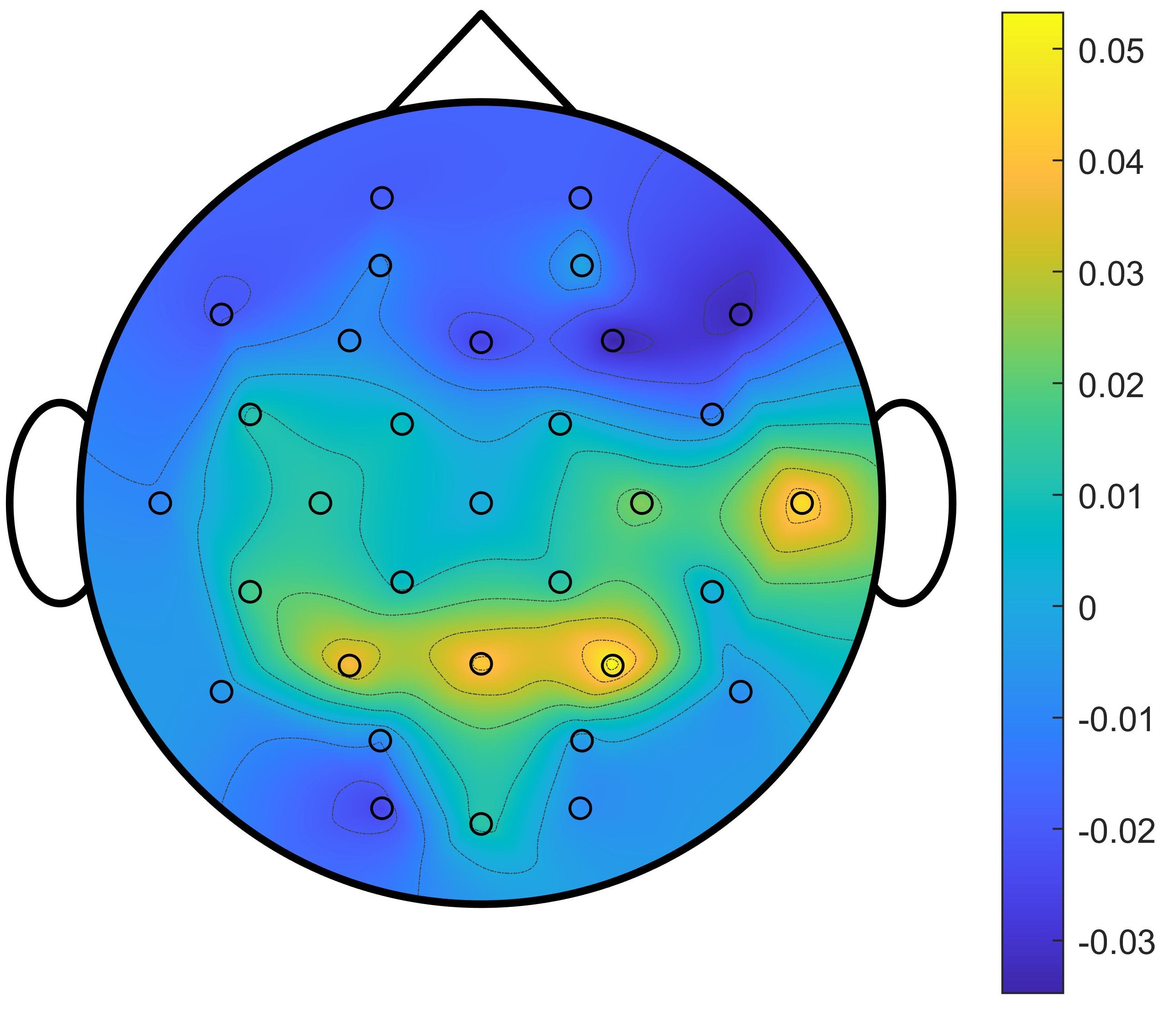}
    \end{minipage}
    \caption{Topographic feature maps for the weight distribution of the first kernel of the last convolutional block of each column in STE during valence classification.The feature values are averaged across the subjects and normalized to the range of $0$ to $1$. (Col: Column)}
    \label{topo1}
\end{figure}
\begin{figure}[t]
    \centering
    \begin{minipage}{0.05\textwidth}
\text{Col. $1$}
\end{minipage}
    \begin{minipage}[c]{.13\textwidth}
     \centering{\small High arousal}
    \includegraphics[width=1\textwidth]{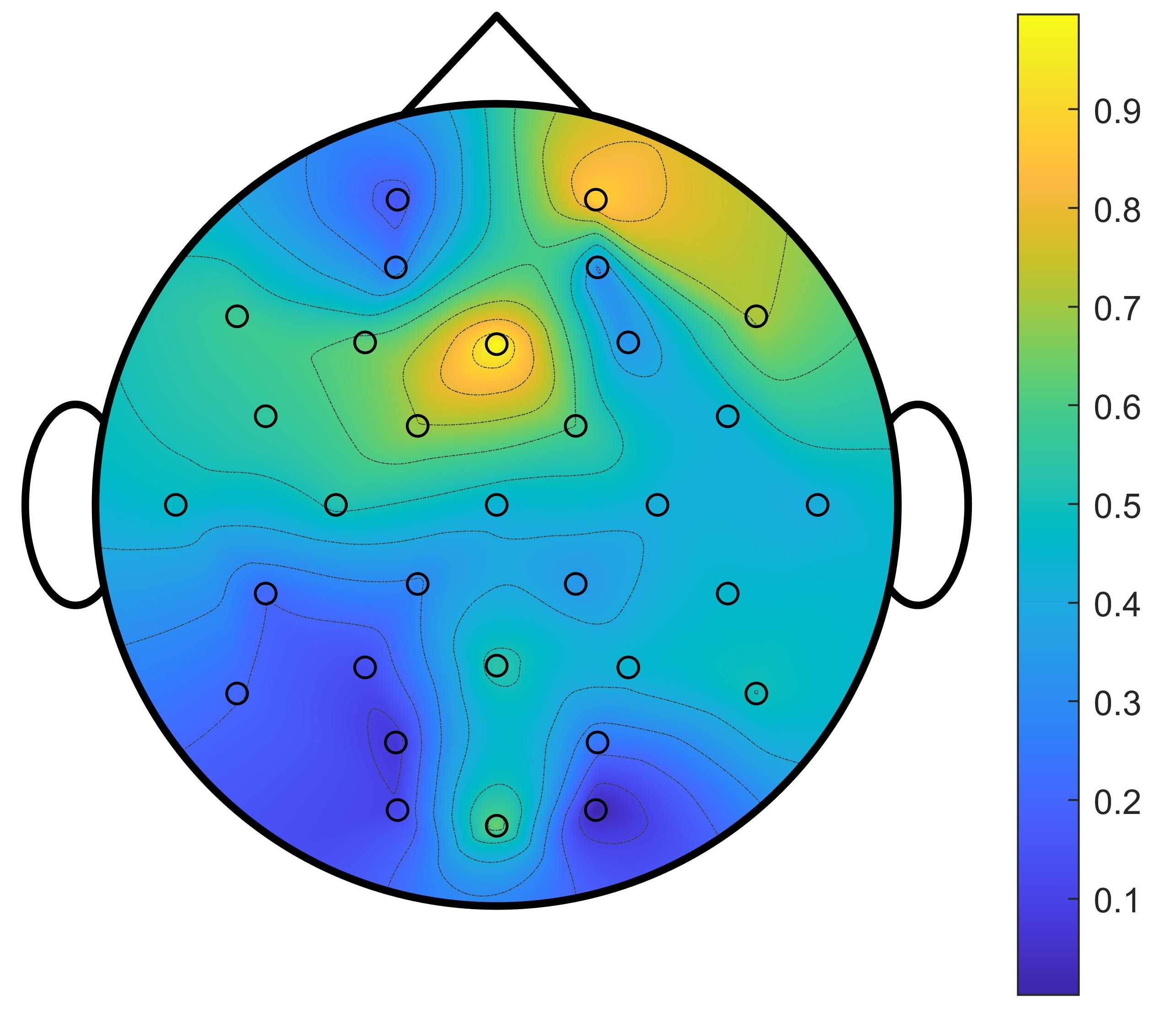}
    \end{minipage}
    \begin{minipage}[c]{.13\textwidth}
    \centering{\small Low arousal}
    \includegraphics[width=1\textwidth]{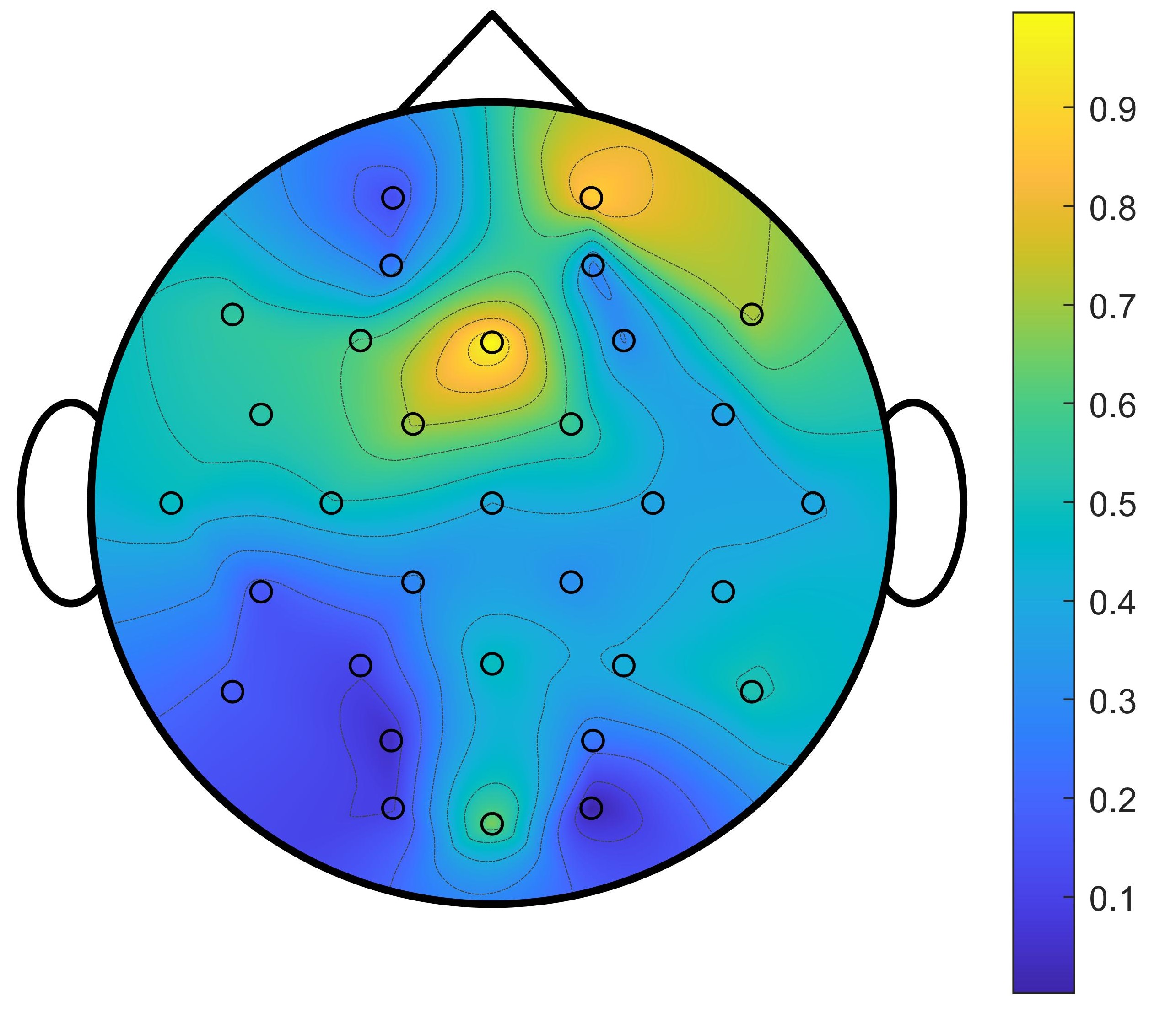}
    \end{minipage}
    \begin{minipage}[c]{.13\textwidth}
    \centering{\small (High-Low) arousal}
    \includegraphics[width=1\textwidth]{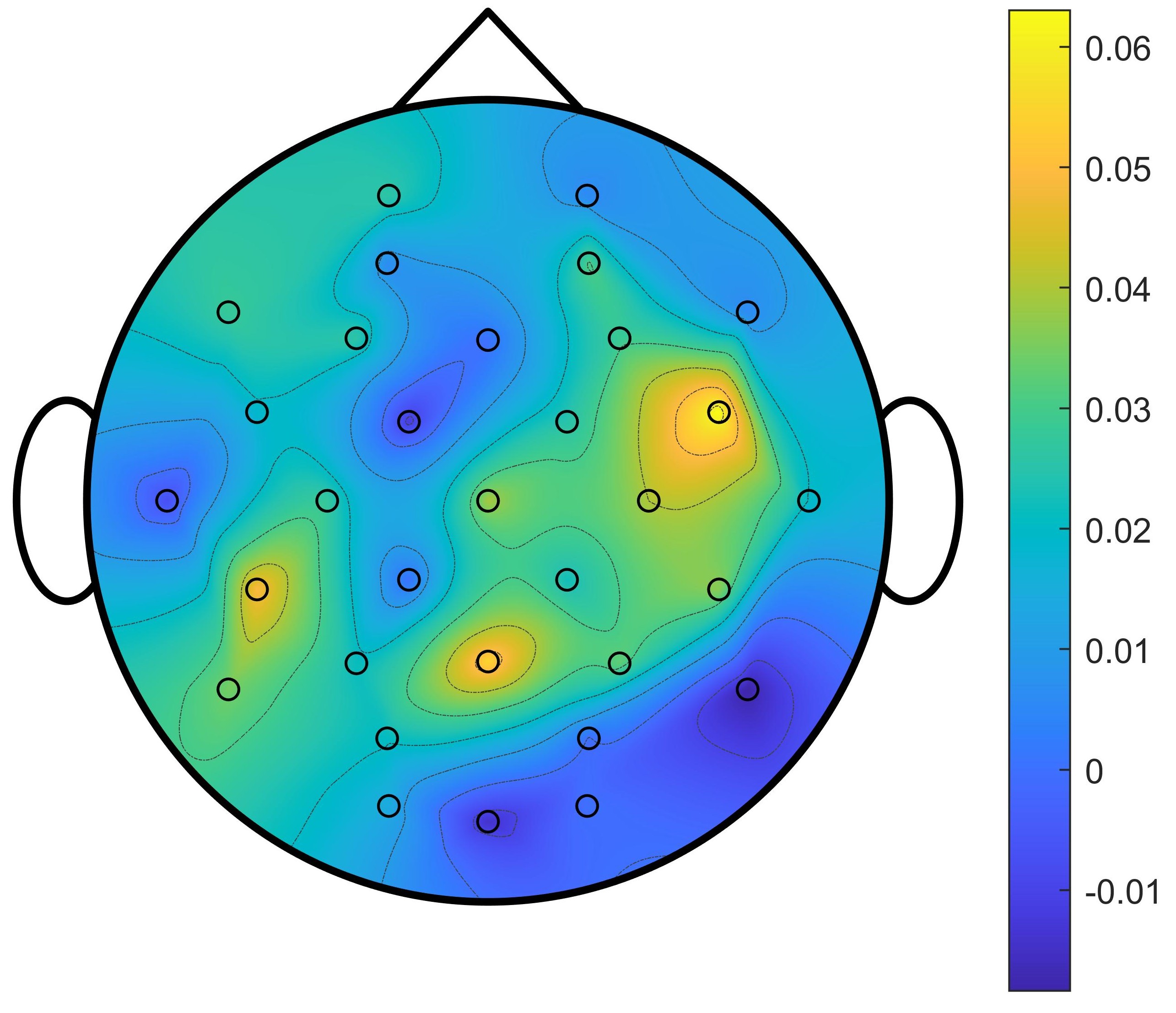}
    \end{minipage}
    \begin{minipage}{0.05\textwidth}
\text{Col. $2$}
\end{minipage}
    \begin{minipage}[c]{.13\textwidth}
    % \centering{\small High arousal}
    \includegraphics[width=1\textwidth]{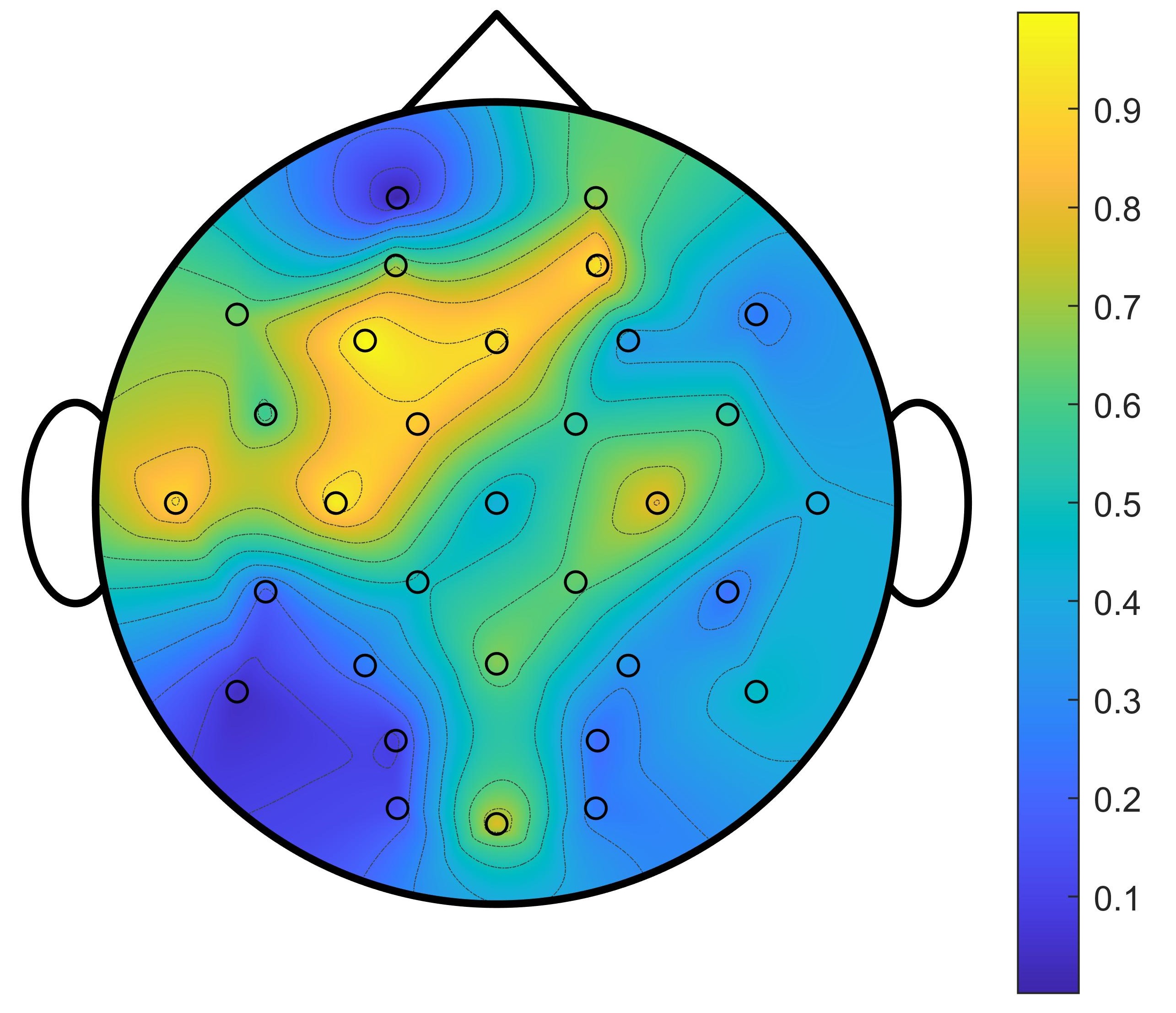}
    \end{minipage}
    \begin{minipage}[c]{.13\textwidth}
    % \centering{\small Low arousal}
    \includegraphics[width=1\textwidth]{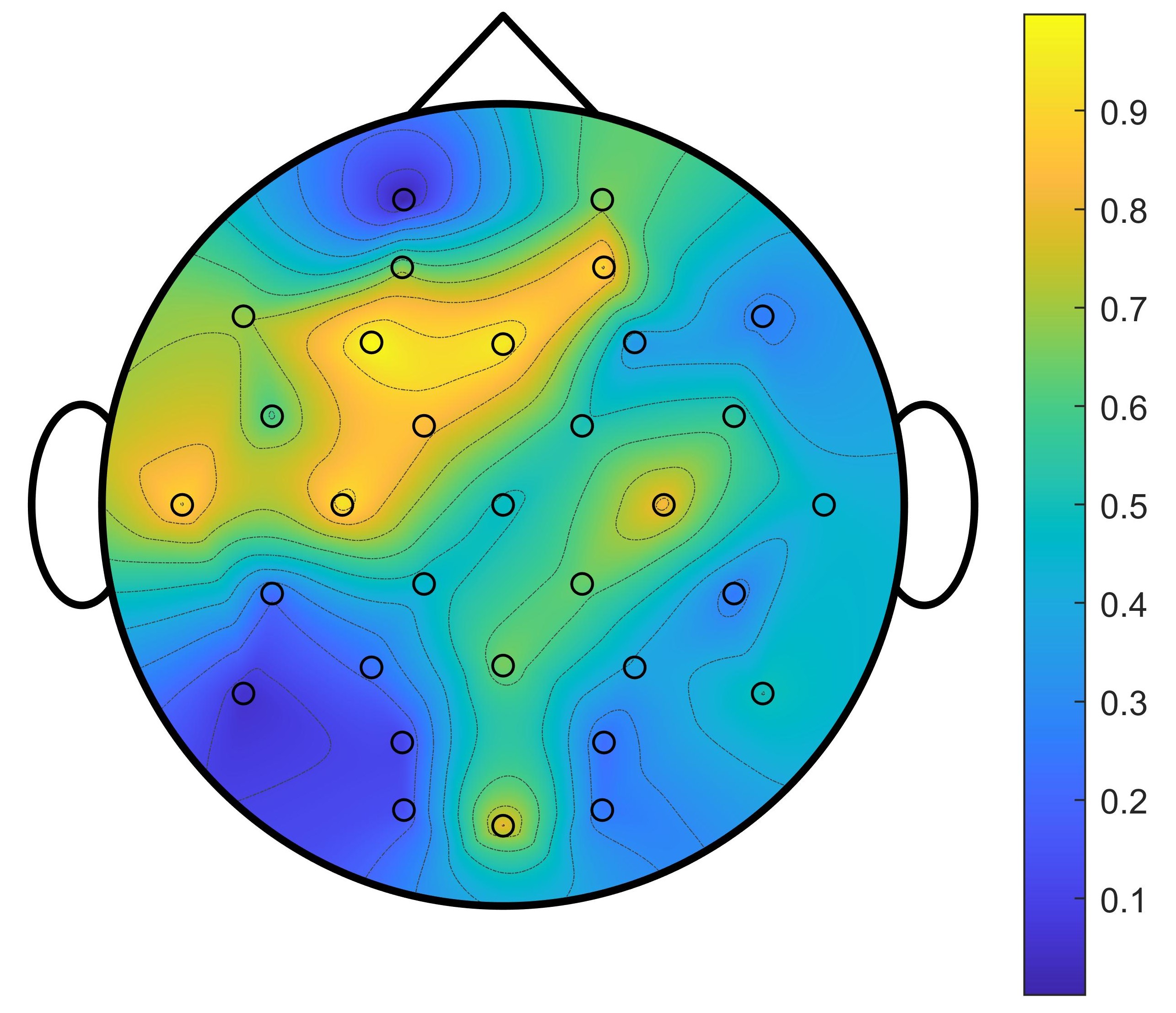}
    \end{minipage}
    \begin{minipage}[c]{.13\textwidth}
    % \centering{\small (High-Low) arousal}
    \includegraphics[width=1\textwidth]{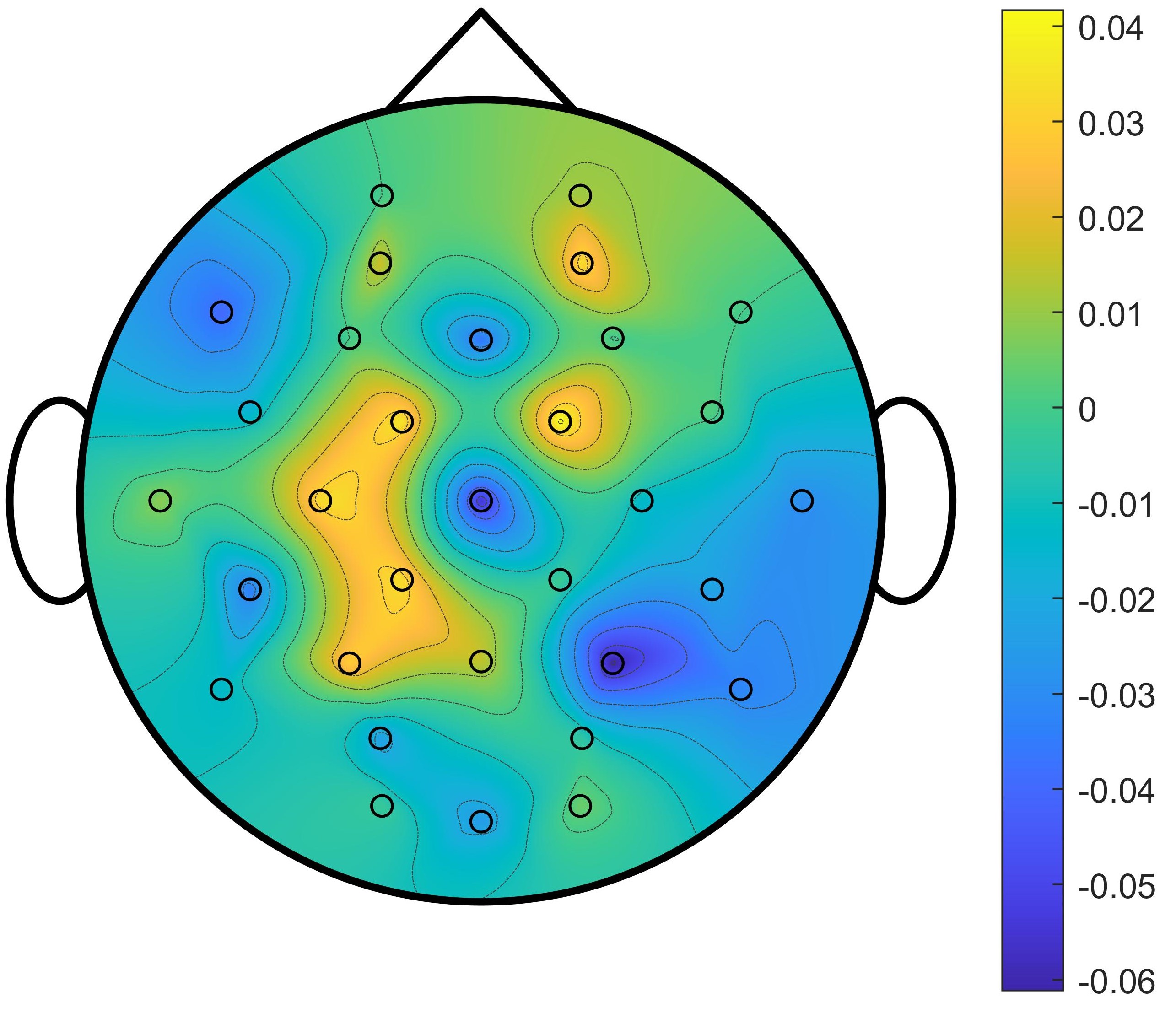}
    \end{minipage}
    \begin{minipage}{0.05\textwidth}
\text{Col. $3$}
\end{minipage}
    \begin{minipage}[c]{.13\textwidth}
    %  \centering{\small High valence}
    \includegraphics[width=1\textwidth]{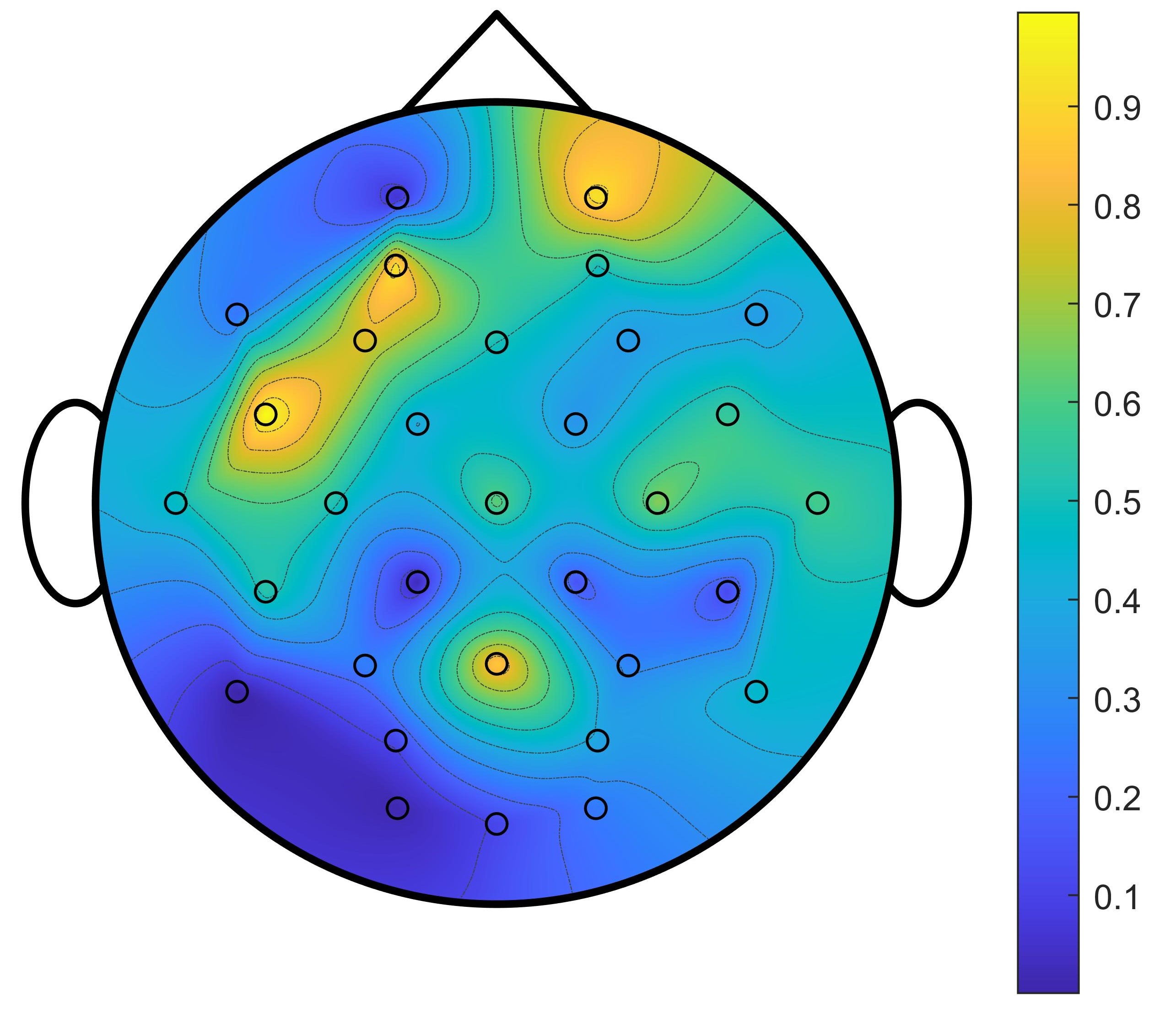}
    \end{minipage}
    \begin{minipage}[c]{.13\textwidth}
    % \centering{\small Low valence}
    \includegraphics[width=1\textwidth]{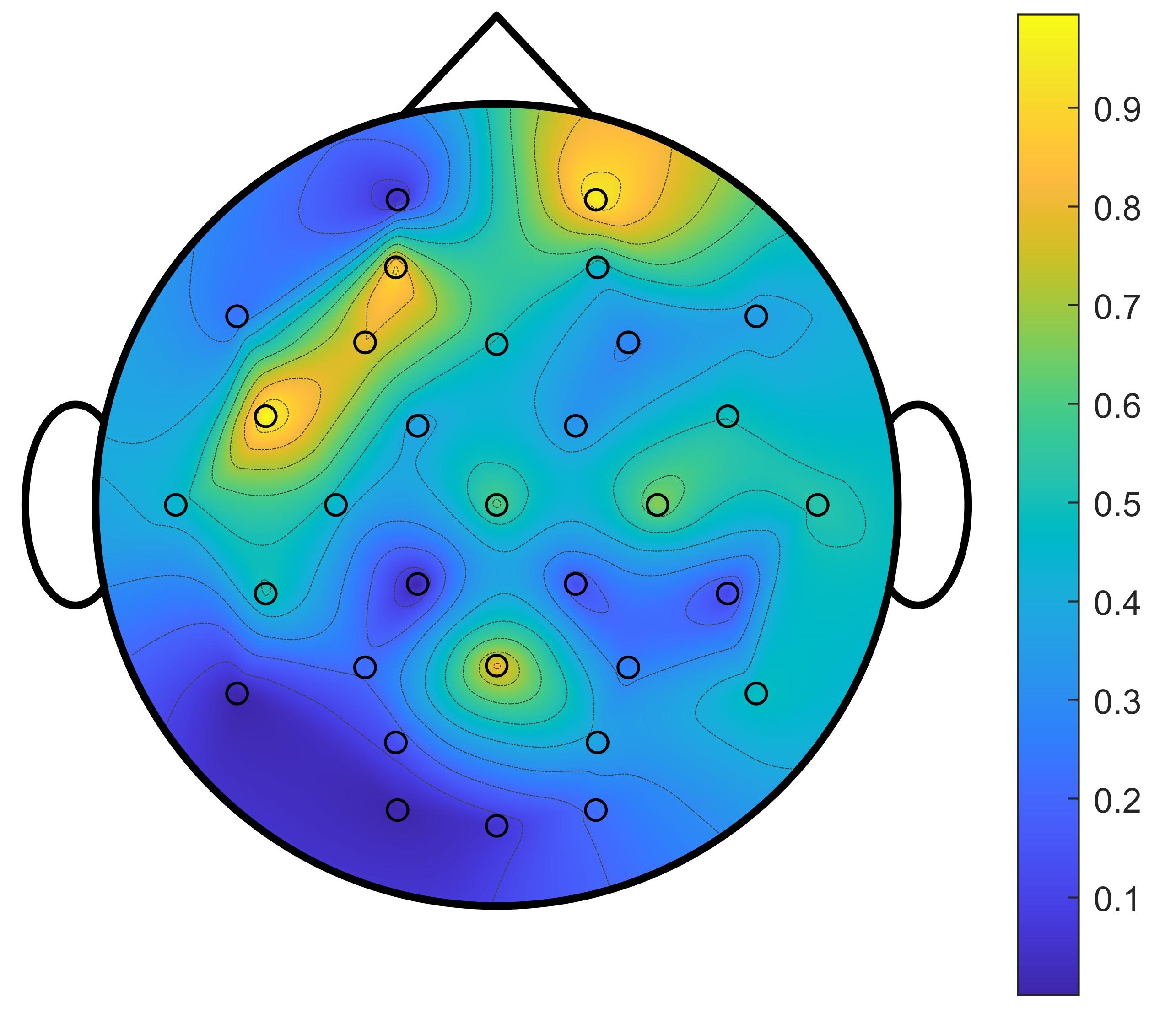}
    % \subcaption{a}
    \end{minipage}
    \begin{minipage}[c]{.13\textwidth}
    % \centering{\small (High-Low) valence}
    \includegraphics[width=1\textwidth]{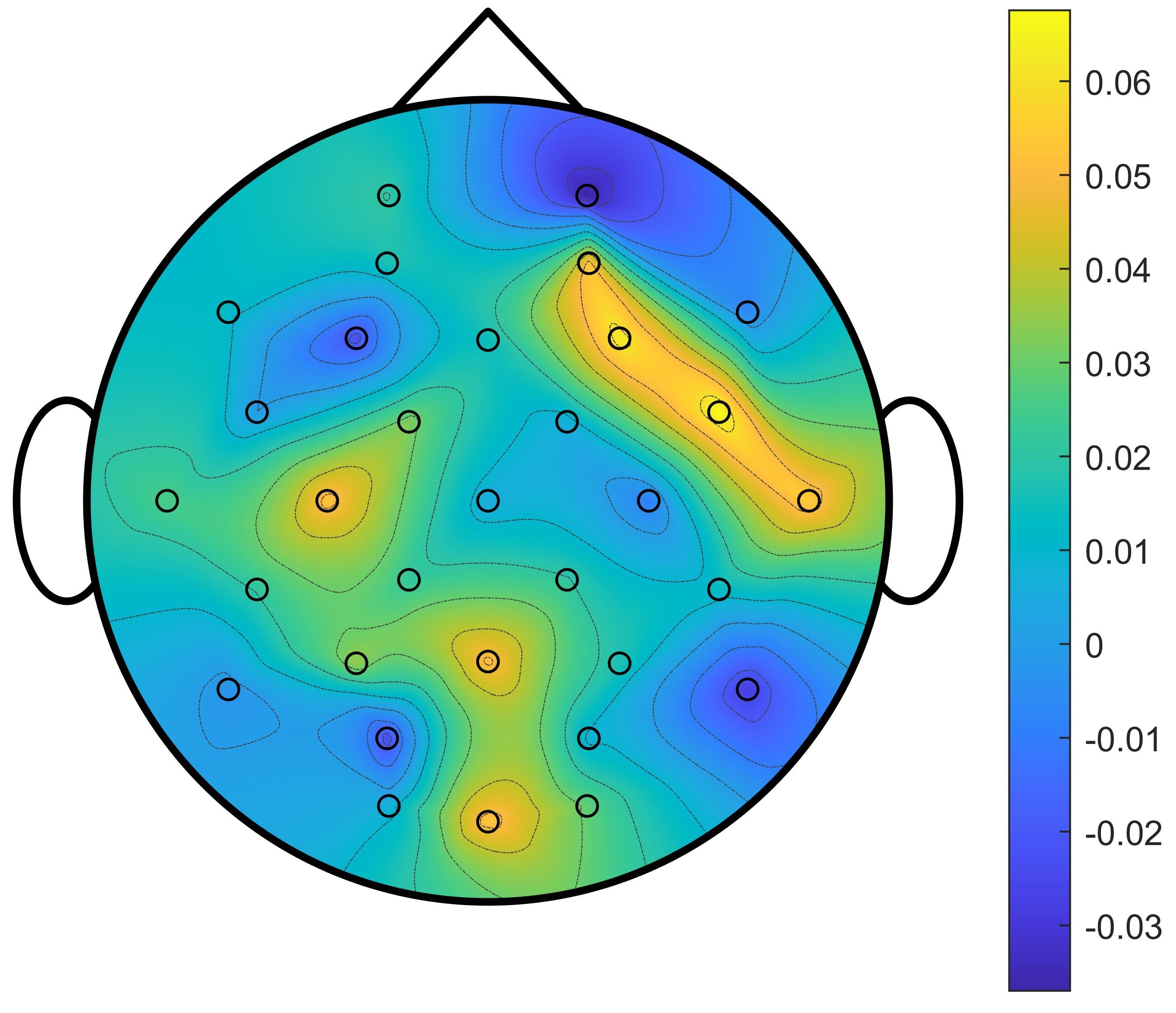}
    \end{minipage}
    \caption{Topographic feature maps for the weight distribution of the first kernel of the last convolutional bock of each column in STE during arousal classification. The feature values are averaged across the subjects and normalized to the range of $0$ to $1$. (Col: Column)}
    \label{topo2}
\end{figure}

% \begin{figure}[t]
%     \centering
   
%     \begin{minipage}[c]{.4\textwidth}
%     % \centering{valence}
%     \includegraphics[width=0.85\textwidth]{s3_val.png}
%     \end{minipage}
%      \begin{minipage}{0.4\textwidth}
%      \centering
% \text{(a)}
% \end{minipage}
%     \begin{minipage}[c]{.4\textwidth}
    
%     \includegraphics[width=0.85\textwidth]{s3_aro.png}
%     \end{minipage}
%     \begin{minipage}{0.4\textwidth}
%  \centering
% \text{(b)}
% \end{minipage}
    
%     % \vspace{-0.01in}
%     \caption{Attention weight values for ($\bf{a}$) valence and ($\bf{b}$) arousal for subject S03 in DEAP dataset. Blue circles, blue bars, and black dots show mean, standard deviation, and data scatter respectively.}
%     \vspace{-0.15in}
%     \label{weight}
% \end{figure}

\begin{table*}[t]
\centering
\caption{Cross-subject TL results for different schemes on valence and arousal classification of DEAP dataset using the proposed SS4-STANN model($DEAP \rightarrow DEAP$).}

\footnotesize{
\begin{tabular}[t]{ l l| c c c c c| c c c c c}
\hline
\multicolumn{2}{c|}{Amount of calibration data ($\mathbf{N}$)} & \multicolumn{5}{c|}{Valence} & \multicolumn{5}{c}{Arousal} \\

\multicolumn{1}{c}{} & \multicolumn{1}{c|}{} &\multicolumn{1}{c}{a} & \multicolumn{1}{c}{b} &  \multicolumn{1}{c}{c} & \multicolumn{1}{c}{d}&
\multicolumn{1}{c|}{e}&\multicolumn{1}{c}{a} & \multicolumn{1}{c}{b} &  \multicolumn{1}{c}{c} & \multicolumn{1}{c}{d}&
\multicolumn{1}{c}{e}\\
\hline
% \midrule
$1$ trial per class & Acc. ($\%$) &$54.1$&$54.1$&$54.2$&$54.3$&$53.3$&$58.3$&$59.0$&$58.5$&$58.6$&$58.7$\\
& F-$1$ score &$0.48$&$0.48$&$0.49$&$0.50$&$0.46$&$0.48$&$0.49$&$0.47$&$0.47$&$0.50$\\
\hline
$10\%$ of data samples & Acc. ($\%$)  &$76.0$&$76.2$&$75.8$&$75.8$&$75.9$&$74.0$&$74.5$&$73.6$&$73.8$&$74.0$\\
& F-$1$ score &$0.73$&$0.74$&$0.73$&$0.74$&$0.74$&$0.70$&$0.71$&$0.69$&$0.70$&$0.70$\\
\hline
$20\%$ of data samples & Acc. ($\%$)  &$83.1$&$82.2$&$82.6$&$82.9$&$83.0$&$81.1$&$81.3$&$80.8$&$80.9$&$81.1$\\
& F-$1$ score &$0.82$&$0.81$&$0.82$&$0.81$&$0.81$&$0.78$&$0.78$&$0.78$&$0.78$&$0.78$\\
\hline
\end{tabular}
}
\label{subjecttosubject}
\end{table*}
\begin{table}[t!]
    \centering
    \caption{Average classification accuracy without transfer learning on the DEAP dataset.}
    \vspace{-0.09in}
    \footnotesize{
    \begin{tabular}[t]{l c  c}
    \hline
\multicolumn{1}{l}{} &\multicolumn{1}{c}{Valence ($\%$)} & \multicolumn{1}{c}{Arousal ($\%$)} \\
        \hline
    %  $N=0$&W-TL               & $0.58$ & $0.59$\\
    % \hline
    $\mathbf{N}=10\%$ of sample data& $72.9$ & $70.9$\\
    \hline
    $\mathbf{N}=20\%$ of sample data& $79.7$ & $78.5$\\
    \hline
    
    \end{tabular}
   
    }
    \label{wodeap}
\end{table}
\begin{table}[t!]
    \centering
    \caption{Average classification accuracy without transfer learning on the DREAMER dataset.}
    \vspace{-0.09in}
    
    \footnotesize{
    \begin{tabular}[t]{l c  c}
    \hline
\multicolumn{1}{l}{} &\multicolumn{1}{c}{Valence ($\%$)} & \multicolumn{1}{c}{Arousal ($\%$)} \\
        \hline
    %  $N=0$&W-TL               & $0.58$ & $0.59$\\
    % \hline
    $\mathbf{N}=10\%$ of sample data& $65.8$ & $73.5$\\
    \hline
    $\mathbf{N}=20\%$ of sample data& $67.6$ & $75.6$\\
    \hline
    $\mathbf{N}=90\%$ of sample data& $70.8$&$77.6$\\
    % &W-TL& $0.83^{*}$&$0.87^{*}$\\
    \hline
    \end{tabular}
    % \medskip
    % \vspace{-0.01in}
    % \newline
    % {$^{*}$ Considering scheme $b$ in tuning process.\par}
    % \vspace{-0.05in}
    }
    \label{table6}
\end{table}
\begin{table*}[t]
\centering
\caption{TL results for different schemes on valence and arousal classification of DREAMER using the proposed SS4-STANN model ($DEAP \rightarrow DREAMER$)}. The classification results for valence and arousal with $\mathbf{N}=0$ (i.e., just based on pre-training) are $58.3\%$ and $59.1\%$, respectively.
\label{table:4}
\footnotesize{
\begin{tabular}[t]{ l l| c c c c c| c c c c c}
\hline
\multicolumn{2}{c|}{Amount of calibration data ($\mathbf{N}$)} & \multicolumn{5}{c|}{Valence} & \multicolumn{5}{c}{Arousal} \\

\multicolumn{1}{c}{} & \multicolumn{1}{c|}{} &\multicolumn{1}{c}{a} & \multicolumn{1}{c}{b} &  \multicolumn{1}{c}{c} & \multicolumn{1}{c}{d}&
\multicolumn{1}{c|}{e}&\multicolumn{1}{c}{a} & \multicolumn{1}{c}{b} &  \multicolumn{1}{c}{c} & \multicolumn{1}{c}{d}&
\multicolumn{1}{c}{e}\\
\hline
% \midrule
$1$ trial per class & Acc. ($\%$) &$61.5$&$64.8$&$63.0$&$63.3$&$62.2$&$64.9$&$68.6$&$67.5$&$69.1$&$65.8$\\
& F-$1$ score &$0.56$&$0.59$&$0.57$&$0.60$&$0.57$&$0.49$&$0.57$&$0.58$&$0.58$&$0.57$\\
\hline
$10\%$ of data samples & Acc. ($\%$)  &$71.1$&$70.4$&$72.0$&$70.9$&$71.9$&$78.1$&$78.2$&$77.9$&$77.9$&$77.4$\\
& F-$1$ score &$0.68$&$0.66$&$0.67$&$0.66$&$0.67$&$0.67$&$0.68$&$0.66$&$0.66$&$0.66$\\
\hline
$20\%$ of data samples & Acc. ($\%$)  &$74.4$&$73.9$&$73.2$&$73.1$&$73.9$&$80.0$&$80.1$&$79.8$&$81.0$&$79.5$\\
& F-$1$ score &$0.71$&$0.70$&$0.69$&$0.69$&$0.70$&$0.71$&$0.71$&$0.72$&$0.72$&$0.69$\\
\hline
$90\%$ of data samples & Acc. ($\%$)  &$82.7$&$83.0$&$79.7$&$81.6$&$82.0$&$83.0$&$87.2$&$84.5$&$85.0$&$86.4$\\
& F-$1$ score &$0.79$&$0.80$&$0.73$&$0.76$&$0.80$&$0.78$&$0.85$&$0.82$&$0.79$&$0.83$\\
\hline
\end{tabular}
}
\label{table7}
\end{table*}

\subsubsection{How does the STE block encode spatial information?} 
To investigate the learning process of the STE block, activations of the last conv layer are visualized to demonstrate wide-band EEG feature maps learned during the valence and arousal recognition tasks. The outputs of the conv layers are averaged along the time dimension to obtain the key spatial features for each kernel separately. This operation results in a $32$-dimensional vector for each kernel. The feature vector is then normalized in the range of $0$ to $1$.

Figures~\ref{topo1} and \ref{topo2} present the topographic scalp plots for the first kernels of each column for valence and arousal, respectively. The representations are averaged over the data samples and subjects. For high-vs-low valence and arousal problems, most of the activities are over frontal, temporal, and central lobes which is consistent with the literature regarding the processing of human emotions \cite{phan2002functional}. The valence difference plots for all columns in Figure~\ref{topo1} show the role of temporal and parietal lobes in positive and negative emotions \cite{nie2011eeg, orgo2015effect}. The difference plot for arousal classification in column $2$ of Figure~\ref{topo2} shows higher activity in the left cortex and frontal lobe similar to previous observations in the literature \cite{harmon2010role}. Moreover, the plots show that arousal processing has a more wide scattered pattern over the brain than valence \cite{siddharth2019utilizing}. 
% Most of the activity is over the frontal
% and central lobes which is consistent with literature regarding the processing of human emotions \cite{rolls1994emotion, phan2002functional}. Moreover, the plots show that arousal processing has a more wide distrusted pattern over the brain than valence.

% \noindent \textbf{How does the RAN block work?} In order to show that the attention mechanism is working properly, we plot the attention weights for a random subject in Figure~\ref{weight}. The output of the attention mechanism is a vector with a size of $1\times{}128$, where $128$ is the number of time steps. The vector elements' values are in the range of $0$ and $1$ and are added up to $1$. As shown in Figure~\ref{weight}, the time steps are given different attention values based on their engagement in task classification. Since each participant could have their own way of thinking and reaction to different stimuli, attention values are scattered in all of the time samples. 
% The trend for valence attention values presents that later time samples got more focus in each data sample. For the arousal case, we see more concentration on early time samples. Since each participant could have his/her way of thinking and reaction to different stimuli, attention values are scattered in all of the time samples.   

\subsection{Evaluating Transfer Learning Performance}
\label{sssec:TLperformance}
To show that the learned representations based on SS-STANN have the generality to be applied in similar tasks that have limited labeled data, we perform TL in two different schemes, namely cross-subject TL and cross-dataset TL. In order to make a fair comparison among the different amounts of calibration data, we set 10\% of the target data samples as a test set, and the calibration data is selected among the rest 90\% of the data. 

\subsubsection{Cross-Subject TL}
To perform cross-subject TL, one subject of the DEAP dataset is considered the target subject and the rest of the subjects are source subjects. Given its superior performance in the supervised classification task, we use SS4-STANN for the TL experiments. First, the network is trained for $100$ epochs on the source subjects' data samples to get the pre-trained network. Second, we use $\mathbf{N}$ data samples of the target subject's data to fine-tune the pre-trained model. During the fine-tuning process, the epoch values are set to $10$ when $\mathbf{N}$ equals $1$ trial per class and $20$ when $\mathbf{N}$ equals $10\%$ and $20\%$ of data samples. Table \ref{subjecttosubject} presents the average TL results on different TL schemes for binary valence and arousal classification. As presented, increasing the number of the calibration data sample improves the classification performance. To validate the TL effect on the classification performance, we present the performance of the proposed model without TL with $\mathbf{N}=10\% ~\text{and}~20\%$ in Table~\ref{wodeap}. Comparing the results of Tables~\ref{subjecttosubject} and~\ref{wodeap}, it is clear that TL helps to increase performance across different subjects.
% Also, the highlighted results indicate that schemes (b) and (a) achieve the best performance almost in all cases which is consistent with the fact that fine-tuning calibrated the detailed features extracted from the last CNN layers.}

\subsubsection{Cross-Dataset TL}
To apply the cross-dataset TL, we trained our proposed model on the whole DEAP dataset and tuned the network on new EEG emotion recognition datasets collected with different stimuli. To this end, we choose the publicly available DREAMER dataset and EEWD. To have consistency between datasets' characteristics, namely EEG electrodes, frequency bands of interest, and the sliding window for segmentation, we choose the following settings. Fourteen common electrodes among all datasets are selected, namely AF3, F7, F3, FC5, T7, P7, O1, O2, P8, T8, FC6, F4, F8, and AF4. The same data slicing process is also applied to the DREAMER dataset. Since the trial length in DREAMER varies from trial to trial, we select the last $60$ s of each trial and segment it into $1$ s data samples. The final number of data samples corresponding to DREAMER is equal to $18\times{}60= 1080$ per participant. The length of each trial sample in EEWD is $1$ s which leaves us with $128$ time steps. For each value of $\mathbf{N}$, the best performing TL scheme is indicated in bold. 

TL process contains two main steps. First, the proposed network is trained on the whole DEAP dataset. Given its superior performance in the supervised classification task, we use SS4-STANN for the transfer learning experiments. $10\%$ of the wide-band EEG samples of each subject in the DEAP dataset is considered for validation and the rest of samples are set aside for training. The network is trained for $100$ epochs and the model parameters corresponding to the least validation loss are considered as the final pre-trained network. Second, we use $\mathbf{N}$ samples of the target data for calibration to fine-tune the pre-trained model. 

In order to have a binary classification problem for the DREAMER dataset, the valence and arousal ratings are divided into two levels using a threshold of $3$. Table \ref{table7} presents the TL results related to the DREAMER dataset for different schemes and various amounts of calibration data. We also report F$1$-score values to avoid the possibility of bias to one class. To avoid overfitting in the fine tuning process for $\mathbf{N}$ equals $1$ trial per class, the number of epochs is set to $1$, and the epoch values is set $20$ when $\mathbf{N}$ equals $10\%,~20\%$, and $90\%$ of data samples. In the latter case, early stopping with a patience parameter of $10$ is set for the tuning process. 
% For each value of $\mathbf{N}$, the best performing TL scheme is indicated in bold. 

As expected, an increase in the amount of calibration data improves the classification performance. To show the effectiveness of transferring pre-trained model parameters, the performance related to $\mathbf{N}=10\%, 20\%, \text{and}~90\%$ of data samples without applying TL is considered as the baseline which is shown in Table \ref{table6}. Observing the results in Tables \ref{table6} and \ref{table7}, it is clear that tuning the pre-trained network increases the classification performance for binary valence and arousal problems. Without using TL, the valence classification accuracy for $\mathbf{N}=10\%, 20\%, \text{and}~90\%$ of data samples are $65.8\%$, $67.6\%$, and $70.8\%$, respectively. However, TL increases the performance for the same $\mathbf{N}$ values to $72.0\%$, $74.4\%$, and $83.0\%$. Similar performance improvement is observed for the arousal classification problem. Considering Tables \ref{table6} and \ref{table7}, for different $\mathbf{N}$ values, the results corresponding for without TL and with TL increase from $73.5\%$, $75.6\%$, and $77.6\%$ to $78.2\%$, $81.0\%$, and $87.2\%$.  
% Also, the results in Table \ref{table7} suggest that even with $1$ labeled trial per class our pre-trained model has the ability to complete the classification task with a performance over the chance level. We observe that by increasing the number of the tuning samples the classification performance also increases.  

To evaluate transfer learning of models trained with EEG in response to video clips to EEG signals obtained in response to written words, we consider the EEWD and perform a binary valence classification scenario since all the words in that experiment are selected from the high arousal group. We consider trials corresponding to rating values lower or equal to $3$ as negative valence and trials correspond to rating values higher or equal to $7$ as positive valence. During the tuning process, the number of epochs and early stopping is set as described in the DREAMER dataset fine-tuning phase. The output prediction of the network is performed on each trial separately. Table \ref{table8} displays the results on the EEWD dataset for different values of $\mathbf{N}$. By considering $\mathbf{N}=1$ trial per class and $\mathbf{N}=10\%, \text{and}~20\%$ of data samples, the performance reaches up to $68.2\%$, $68.9\%$, and $73.5\%$ which is related to schemes a, c, and b, respectively. Due to the limited amount of data samples we do not present the results similar to Table \ref{table6} for this dataset since the model overfits in early epochs. 
\begin{table}[t]
\centering
\caption{TL results for different schemes on valence classification of Emotional English Word Dataset (EEWD) ($DEAP \rightarrow EEWD$)}.

\footnotesize{
\begin{tabular}[t]{ l l| c c c c c}
\hline
\multicolumn{2}{c|}{Amount of calibration data ($\mathbf{N}$)} & \multicolumn{5}{c}{Valence} \\

\multicolumn{1}{c}{} & \multicolumn{1}{c|}{} &\multicolumn{1}{c}{a} & \multicolumn{1}{c}{b} &  \multicolumn{1}{c}{c} & \multicolumn{1}{c}{d}&
\multicolumn{1}{c}{e}\\
\hline
% \midrule
$1$ trial per class & Acc. ($\%$)  &$68.2$&$66.5$&$64.5$&$67.3$&$65.0$\\
& F-$1$ score &$0.68$&$0.66$&$0.65$&$0.64$&$0.64$\\
\hline
$10\%$ of data samples & Acc. ($\%$)  &$67.0$&$67.7$&$68.9$&$66.9$&$67.0$\\
& F-$1$ score &$0.66$&$0.67$&$0.68$&$0.66$&$0.67$\\
\hline
$20\%$ of data samples & Acc. ($\%$)  &$73.0$&$73.5$&$70.6$&$70.2$&$70.9$\\
& F-$1$ score &$0.71$&$0.73$&$0.69$&$0.69$&$0.70$\\
\hline
\end{tabular}
}
\label{table8}
\end{table}
\subsubsection{Verification of TL}
\begin{figure}[t]
    \centering
    \begin{minipage}{0.05\textwidth}
\text{\small{Valence}}
\end{minipage}
    \begin{minipage}[c]{.18\textwidth}
    % \centering{valence}
    \includegraphics[width=0.9\textwidth]{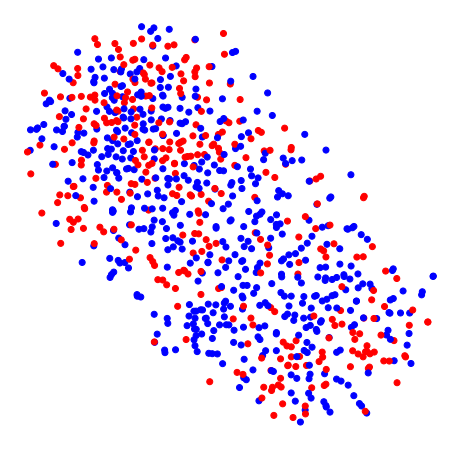}
    \end{minipage}
    \begin{minipage}[c]{.2\textwidth}
    
    \includegraphics[width=0.9\textwidth]{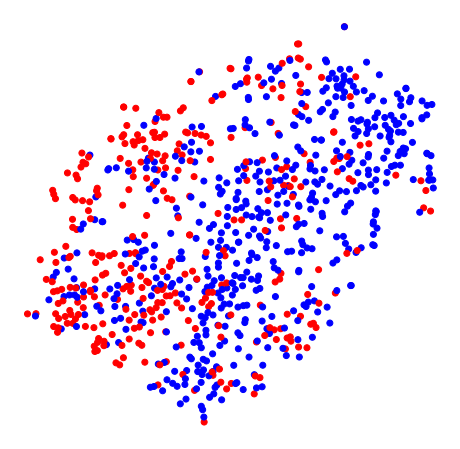}
    \end{minipage}
    \begin{minipage}{0.05\textwidth}
\text{\small{Arousal}}
\end{minipage}
    \begin{minipage}[c]{.18\textwidth}
    % \centering
    \includegraphics[width=0.9\textwidth]{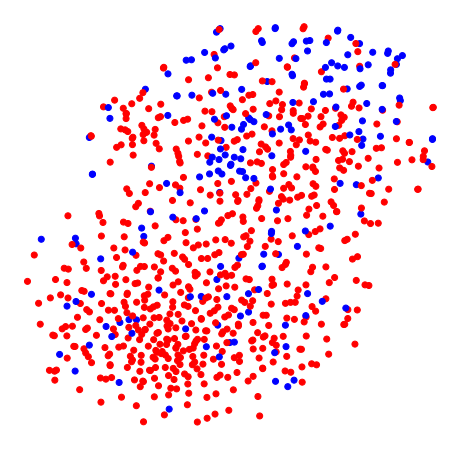}
    
    \end{minipage}
    \begin{minipage}[c]{.2\textwidth}
    % \centering{arousal}
    \includegraphics[width=0.9\textwidth]{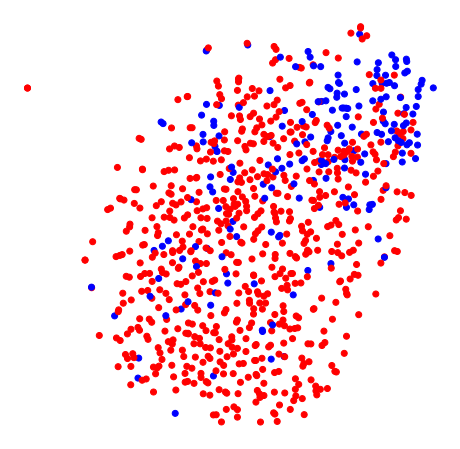}
    
    \end{minipage}
    \begin{minipage}{0.18\textwidth}
 \centering
\text{(a)}
\end{minipage}
\begin{minipage}{0.18\textwidth}
 \centering
\text{(b)}
\end{minipage}
    % \vspace{-0.01in}
    \caption{Scatter plot of representative samples in the last dense layer ahead of the classification softmax layer, visualized using t-SNE for valence (first row) and arousal (second row). (\textbf{a}) before TL and fine-tuning, and (\textbf{b}) after TL and fine tuning with $\mathbf{N}=20\%$. Red and blue dots present the data sample for each class.}
    \vspace{-0.15in}
    \label{tsne}
\end{figure}

To establish that the proposed structure works properly and TL helps to separate the components related to each class, we randomly pick the EEG samples of one of the subjects in DREAMER dataset to visualize them with t-SNE \cite{vazquez2013virtual} (Figure~\ref{tsne}). Figure~\ref{tsne}(a) shows the scatter plot of samples in the last dense layer before the classification softmax layer without TL and fine-tuning. Figure~\ref{tsne}(b) presents the results after TL and fine-tuning. As can be seen, before fine-tuning the process, the representations corresponding to different classes are more mixed up and TL helps make them more separable.  
\section{Conclusion}
\label{sec:conclusion}
In this study, we proposed a novel deep learning architecture for subject-dependent EEG-based emotion classification tasks. The proposed SS-STANN involves a hybrid structure with parallel STE and RAN blocks and an attention mechanism. SS-STANN captures the spatial and temporal information inherent in multi-channel EEG data while enforcing graph smoothness in the spatial domain. We demonstrated that this work performs better than other state-of-the-art solutions, with classification accuracies of over 95.0$\%$ for valence, arousal, and dominance for the DEAP dataset. Moreover, the critical frequency bands and regions are explored to validate the performance results. We also showed that the representations extracted from one EEG experiment could be used in other EEG emotion recognition tasks with similar and different stimulus modalities, highlighting the cross-modal transferability potential of the trained model and learned representations. In the future, we will investigate the effect of using different modalities along with EEG in a similar problem. Also, we will concentrate on conducting real-time experiments based on the proposed framework. Moreover, future work may involve adopting the spatio-temporal feature learning ideas presented here to problems involving other modalities of physiological data such as functional MRI (fMRI) or magnetoencephalography (MEG).

\bibliographystyle{IEEEbibvv}

\bibliography{refs}
\newpage
\begin{IEEEbiography}[{\includegraphics[width=1in,height=1.25in,clip,keepaspectratio]{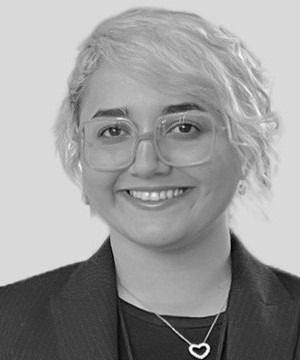}}]{Shadi Sartipi}
is currently pursuing a Ph.D. degree in the
Electrical and Computer Engineering Department at the University of
Rochester, Rochester, NY. 

Her research interests include brain-computer interfaces (BCI), affective computing, sleep scoring, machine learning, and signal processing of physiological data. Shadi received the IEEE Brain Best Paper Award in 2021.
\end{IEEEbiography}
\begin{IEEEbiography}[{\includegraphics[width=1in,height=1.25in,clip,keepaspectratio]{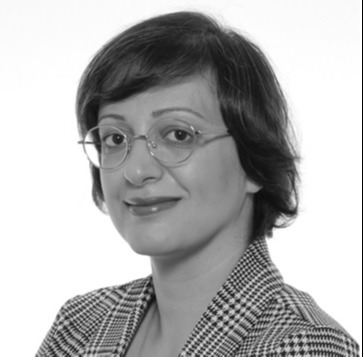}}]{Mastaneh Torkamani-Azar}
received her Ph.D. from Sabanci University, Turkey, in 2020. As a Research Assistant at Signal Processing and Information Systems laboratory, she developed statistical signal processing techniques for EEG-based vigilance inference and adaptive brain-computer interfaces. She was a Postdoctoral Researcher with School of Computing, University of Eastern Finland, specializing in multi-modal signal processing for surgical skill assessment. She led a proof-of-concept project on expanded reality for minimally invasive surgery that secured research-to-business funding from Business Finland in 2023. She is currently a Postdoctoral Researcher at A. I. Virtanen Institute for Molecular Sciences, University of Eastern Finland, focusing on cross-modality brain mapping and data provenance for multiscale assessment of epileptogenicity. 

Her research interests include biomedical signal and image processing, cognitive neuroscience, affective computing, adaptive BCIs, and interactive medical technologies. 
\end{IEEEbiography}
\begin{IEEEbiography}[{\includegraphics[width=1in,height=1.25in,clip,keepaspectratio]{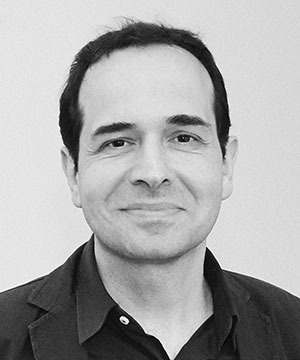}}]{Mujdat Cetin}
is a Professor of Electrical and Computer Engineering and the Director of the Goergen Institute for Data Science at the University of Rochester. 

His research interests include computational imaging, bioimage analysis, and brain-computer/machine interfaces. Prof. Cetin is currently the Editor-in-Chief of the IEEE Transactions on Computational Imaging, a Senior Area Editor for the IEEE Transactions on Image Processing, as well as an Associate Editor for the SIAM Journal on Imaging Sciences and for Data Science in Science.
He served as the Chair of the IEEE Computational Imaging Technical Committee and as the Technical Program Co-chair for five conferences. Prof. Cetin has received several awards including the IEEE Signal Processing Society Best Paper Award; the EURASIP/Elsevier Signal Processing Best Paper Award; the IET Radar, Sonar and Navigation Premium Award; and the Turkish Academy of Sciences Distinguished Young Scientist Award. He is a Fellow of IEEE.
\end{IEEEbiography}
% \vfill

\end{document}